\documentclass[times,onecolumn,review,nopreprintline]{elsarticle}

\usepackage{times}
\usepackage{epsfig}
\usepackage{graphicx}
\usepackage{amsmath}
\usepackage{amssymb}
\usepackage{booktabs}
\usepackage{multirow}
\usepackage{array}
\usepackage{subfig}
\usepackage{bm}
\usepackage{yjvci}
\usepackage{framed,multirow}
\usepackage[flushleft]{threeparttable}

\usepackage{amssymb}
\usepackage{latexsym}

\usepackage{algorithm}
\usepackage{algorithmic}
\usepackage{lineno}

\usepackage{url}
\usepackage{xcolor}

\usepackage{hyperref}

\definecolor{newcolor}{rgb}{.8,.349,.1}

\makeatletter
\newif\if@restonecol
\makeatother

\begin{document}


\begin{frontmatter}

\title{DDR-ID: Dual Deep Reconstruction Networks Based Image Decomposition for Anomaly Detection}%

\author[1]{Dongyun \snm{Lin}\corref{cor1}}
\ead{lin_dongyun@i2r.a-star.edu.sg}
\cortext[cor1]{Corresponding author:
  Tel.: +6584342859;}
\author[1]{Yiqun \snm{Li}}
\ead{yqli@i2r.a-star.edu.sg}
\author[1]{Shudong \snm{Xie}}
\ead{xie_shudong@i2r.a-star.edu.sg}
\author[1]{Tin Lay \snm{Nwe}}
\ead{tlnma@i2r.a-star.edu.sg}

\author[1]{Sheng \snm{Dong}}
\ead{dong_sheng@i2r.a-star.edu.sg}

\address[1]{Institute for Infocomm Research, A*STAR, Singapore, 1 Fusionopolis Way, $\#$21-01 Connexis (South Tower), Singapore 138632}


\begin{abstract}

One pivot challenge for image anomaly (AD) detection is to learn discriminative information only from normal class training images. Most image reconstruction based AD methods rely on the discriminative capability of reconstruction error. This is heuristic as image reconstruction is unsupervised without incorporating normal-class-specific information. In this paper, we propose an AD method called dual deep reconstruction networks based image decomposition (DDR-ID). The networks are trained by jointly optimizing for three losses: the one-class loss, the latent space constrain loss and the reconstruction loss. After training, DDR-ID can decompose an unseen image into its normal class and the residual components, respectively.  Two anomaly scores are calculated to quantify the anomalous degree of the image in either normal class latent space or reconstruction image space. Thereby, anomaly detection can be performed via thresholding the anomaly score. The experiments demonstrate that DDR-ID outperforms multiple related benchmarking methods in image anomaly detection using MNIST, CIFAR-10 and Endosome datasets and adversarial attack detection using GTSRB dataset.
\end{abstract}


\end{frontmatter}


\section{Introduction}
\label{Introduction}
In computer vision, anomaly detection (AD) is a one-class classification task of predicting an image as the normal class (inliers) or the anomalous classes (outliers)~\cite{chandola2009anomaly,pimentel2014review}. The training process of an AD method only exploits the normal class training images. Since the distribution of anomalous classes cannot be empirically studied using training samples, AD is treated as an unsupervised learning problem. For testing, the method is required to quantify the anomalous degree of an unseen testing image whose ground truth label could be either normal or anomalous. This asymmetrical setting in training and testing is very common in real-world applications, e.g., video surveillance for anomalous events~\cite{dos2019generalization,afiq2019review}, medical image diagnosis for malignant areas~\cite{seebock2016identifying,schlegl2017unsupervised} and remote sensing image analysis of anomalous signals~\cite{matteoli2014overview,li2017transferred}. Hence, one pivot challenge to address an AD task is to learn discriminative representations only from normal class images.

Several pioneering statistical learning methods, such as One-Class Support Vector Machine (OC-SVM)~\cite{Scholkopf:2001:ESH:1119748.1119749}, One-Class Support Vector Data Description (OC-SVDD)~\cite{Tax2004} and Kernel Density Estimation (KDE)~\cite{10.2307/2237880}, were proposed to address image related AD tasks relying on substantial manual engineering of image features and kernel parameters. These methods generalize well in applications with small data but suffer from poor scalability to large data scenarios. Recently, deep neural network based methods achieve great success in multiple vision tasks, e.g., image classification~\cite{krizhevsky2012imagenet,szegedy2015going,he2016deep}, detection~\cite{ren2015faster,liu2016ssd} and segmentation~\cite{long2015fully,ronneberger2015u}, mainly benefiting from that they provide a way to learn discriminative representations of images in an end-to-end manner via deep architectures. Several deep neural network based methods have been proposed to handle image related AD problems~\cite{chen2017outlier,amarbayasgalan2018unsupervised,sakurada2014anomaly,schlegl2017unsupervised,sabokrou2018adversarially,ruff2018deep}. Among them, some exploit deep neural network models, e.g., Deep Convolution Autoencoder (DCAE) or Generative Adversarial Network (GAN), to reconstruct normal class images and assume the models could produce lower reconstruction errors for normal class images than those anomalous ones~\cite{chen2017outlier,amarbayasgalan2018unsupervised,sakurada2014anomaly,schlegl2017unsupervised}. This assumption is somehow heuristic as the training for reconstruction is conducted unsupervisedly without considering normal-class-specific information. Intuitively, a perfect autoencoder performs an image-to-image identity mapping and therefore produces zero reconstruction error for both normal class and anomalous class images. In this case, the reconstruction error is not discriminative for anomaly detection. To alleviate this issue, several recent AD methods achieved promising performance by exploiting end-to-end learning processes with AD related optimization objectives~\cite{sabokrou2018adversarially,ruff2018deep}. We will briefly review the related works in Section \ref{RelatedWorks}.


\begin{figure}[!t]
   \centering
   \includegraphics[scale=0.4]{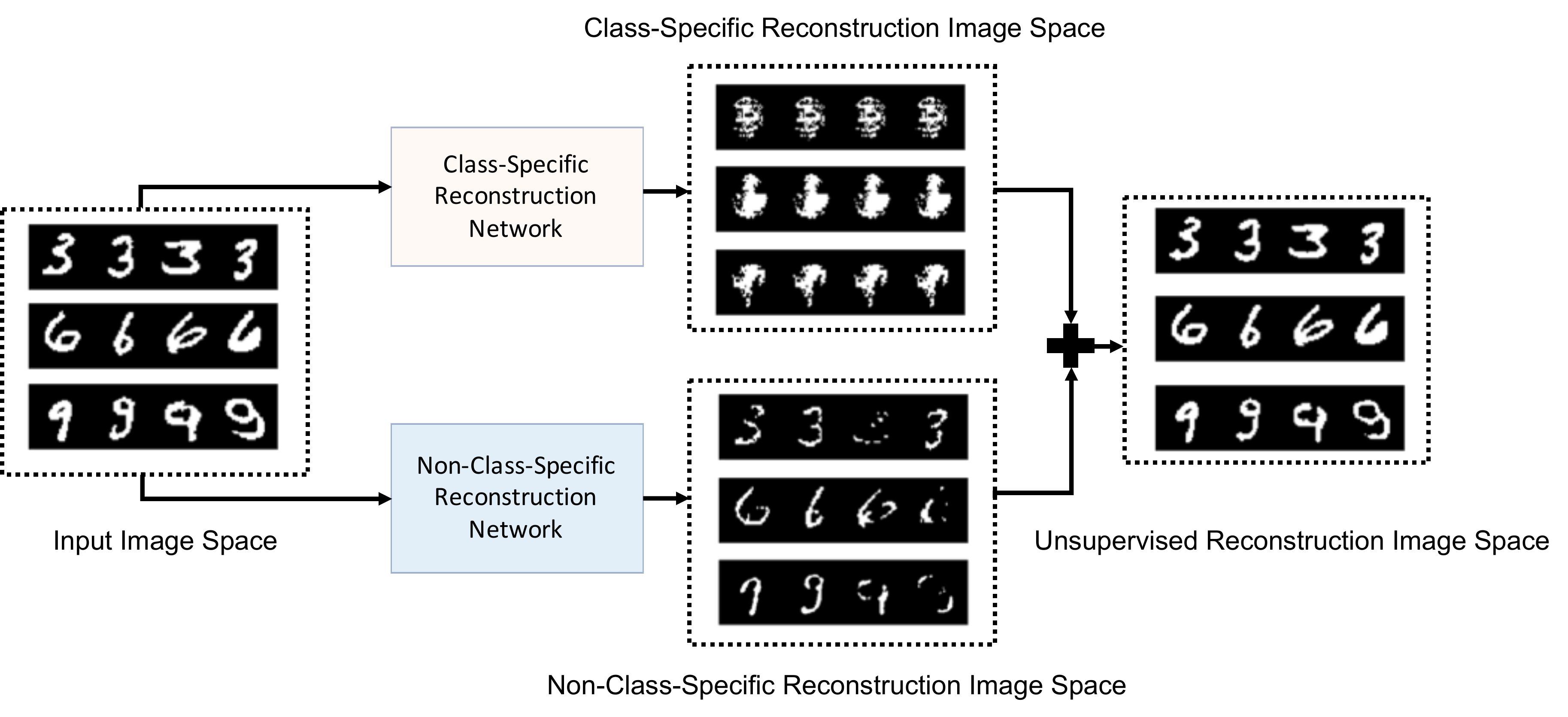}
   \caption{An illustration of the image decomposition process of our DDR-ID using MNIST digits. \textbf{The upper path}: class-specific component extraction. \textbf{The lower path}: non-class-specific residual component extraction.}
\label{IntuitiveIllustration}
\end{figure}
{\color{black}{In this paper, we propose an effective image decomposition method for anomaly detection based on dual deep reconstruction networks (DDR-ID).}} The proposed method aims to achieve normal-class-specific image decomposition in an end-to-end manner by optimizing for AD oriented objectives together with image reconstruction. We train the image decomposition network using only the normal class images. This network divides the unsupervised image reconstruction into the image decomposition stage followed by the superposition stage. The former decomposes an image into its class-specific component and non-class-specific component, respectively, by simultaneously feeding the image through two encoder-decoder reconstruction networks. The latter adds these two components to reconstruct the original input image. Here, in the AD context, class-specific and non-class-specific components refer to normal class component and non-normal class component, respectively. Fig.~\ref{IntuitiveIllustration} illustrates the image decomposition process using MNIST digit images~\cite{lecun2010mnist}. After decomposition, we build ``normal class templates'', i.e., the mean representations in both normal class latent vector space and reconstructed image space, using the normal class components extracted from the training images. For inference, given an unseen image, its normal class component is extracted by feeding the image through the class-specific reconstruction network (the upper path of Fig.~\ref{IntuitiveIllustration}) and its anomaly is quantified by the difference between its normal class component and the precomputed ``normal class templates''.


The major contributions of this paper are summarized as follows: {\color{black}{(i) we propose a dual deep reconstruction networks based image decomposition (DDR-ID) which can alleviate the limitation of unsupervised reconstruction based AD methods by extracting the normal class component from the input image for robust anomaly detection while the non-normal-class component is filtered out. The proposed method exploits an end-to-end training process by jointly optimizing for three losses: the one-class loss, the auxiliary latent space constrain loss and the reconstruction loss;}} (ii) we propose two anomaly scores utilizing the normal class reconstruction network to effectively quantify the anomalous degree of a testing image. The prediction of anomalous images can be performed by thresholding the corresponding anomaly scores; (iii) the proposed DDR-ID outperforms multiple related benchmarking methods in image anomaly detection using MNIST, CIFAR-10 and Endosome datasets and adversarial attack detection using GTSRB dataset.





\section{Related Works}
\label{RelatedWorks}

In this section, we mainly review two types of AD methods, namely image reconstruction based methods and AD oriented objective based methods. Deep autoencoder (AE) is a common unsupervised network for image reconstruction. An AE consists of an encoder followed by a decoder. It first exploits the encoder to map the input image into a latent space vector and then the decoder to reconstruct the image. Multiple AE variants, such as denoising autoencoder~\cite{sabokrou2018adversarially}, variational autoencoder (VAE)~\cite{an2015variational}, convolution autoencoder (CAE)~\cite{sabokrou2018deep,Xia_2015_ICCV}, have been extensively explored to address AD problems. Most existing AE based methods are built upon the observations that reconstruction error is discriminative, i.e., an autoencoder trained using only normal class images should produce smaller reconstruction errors for normal class images than those anomalous ones~\cite{Xia_2015_ICCV}. Hence, the reconstruction error is often used as a metric to describe the anomalous degree of testing images. However, the discriminative capability of the reconstruction error produced by AEs is somehow unclear since the training of an AE aims to minimize the reconstruction error rather than AD oriented losses~\cite{ruff2018deep}.



Apart from deep AEs which construct deep neural networks to enable image-to-image mapping, Generative Adversarial Networks (GANs)~\cite{goodfellow2014generative} is formulated to generate images directly from an input vector space. A GAN model consists of a generator network $G$ and a discriminator network $D$ trained using an adversarial strategy: (i) the generator network $G$ is trained to generate realistic image data which can ``fool'' the discriminator $D$; (ii) the discriminator network $D$ is trained to discriminate the ``real'' images from the ``fake'' images generated by $G$. After training, $G$ performs a vector-to-image mapping which is similar to the decoder of an AE. There is a surge of research works in applying GAN to address AD problems~\cite{schlegl2017unsupervised,sabokrou2018adversarially}. Ref. \cite{schlegl2017unsupervised} proposes Anomaly detection with GAN (AnoGAN) to detect disease markers from medical images. AnoGAN first trains a GAN model using the normal class images. Given an unseen testing image, AnoGAN calculates the optimal latent vector which can best reconstruct the testing image. This process is performed by jointly minimizing a residual loss and a discriminant loss. The anomaly degree of the testing image is calculated by the value of the overall loss function when the training converges. {\color{black}{Similar to AnoGAN, ADGAN~\cite{deecke2018image} calculates the optimal latent vector only considering the residual loss along with a small amount of parameter tuning of the generator.}} Adversarially Learned One-Class Classifier (ALOCC) \cite{sabokrou2018adversarially} uses the adversarial learning strategy similar to GAN to train a denoise autoencoder $\mathcal{R}$ and a discriminator $\mathcal{D}$. $\mathcal{R}$ is trained to map a noised normal class image into the image that $\mathcal{D}$ cannot identify while $\mathcal{D}$ is trained to discriminate clean normal class images from denoised images reconstructed by $\mathcal{R}$. ALOCC assumes that the denoising process is discriminative thanks to the supervision of discriminator $\mathcal{D}$.



Both AE and GAN based AD methods are built upon the assumption that the reconstruction process is discriminative in classifying normal class and anomalous class images. However, the image reconstruction based methods do not directly optimize for AD oriented objectives. It is usually tricky to formulate an unsupervised objective for end-to-end learning~\cite{bengio2013representation} and there are some promising attempts such as deep unsupervised clustering~\cite{xie2016unsupervised}. Ref. \cite{ruff2018deep} proposed a method called Deep Support Vector Data Description (Deep SVDD) which exploits the idea of kernel-based one-class SVM using deep neural networks. Specifically, Deep SVDD replaces the manual feature and kernel engineering by a deep neural network architecture and trains the network to minimize the volume of a hypersphere that encloses the latent representations of the normal class images. The anomalous degree of an unseen image can be quantified by the distance between the latent representation vector of the image and the sphere centroid. {\color{black}{One-class neural network (OC-NN)~\cite{chalapathy2018anomaly} adopts a one-class SVM-like loss to train a feed forward network with one hidden layer to generate a hyperplane to separate all the normal class data from the origin. The anomalous degree of an image is therefore quantified by the distance to the origin.}} In this paper, we mainly investigate to effectively incorporate AD oriented objectives into the image reconstruction process to alleviate the aforementioned limitations met by most AE and GAN based AD methods.



\section{Proposed Method}

We propose an anomaly detection method which exploits an image decomposition architecture based on a pair of deep reconstruction neural networks, namely the class-specific reconstruction network $\mathcal{R}_C$ and the non-class-specific reconstruction network $\mathcal{R}_N$. $\mathcal{R}_C$ is proposed to extract normal class information, i.e., the commonality among images generated from the normal class distribution. $\mathcal{R}_N$, on the other hand, acts as the residual counterpart of $\mathcal{R}_C$ since it extracts non-normal class information. Both $\mathcal{R}_C$ and $\mathcal{R}_N$ have the encoder-decoder architecture. Apart from $\mathcal{R}_C$ and $\mathcal{R}_N$, we also propose an auxiliary discriminator network to regularize the latent space of $\mathcal{R}_C$.

\begin{figure}[hptb]
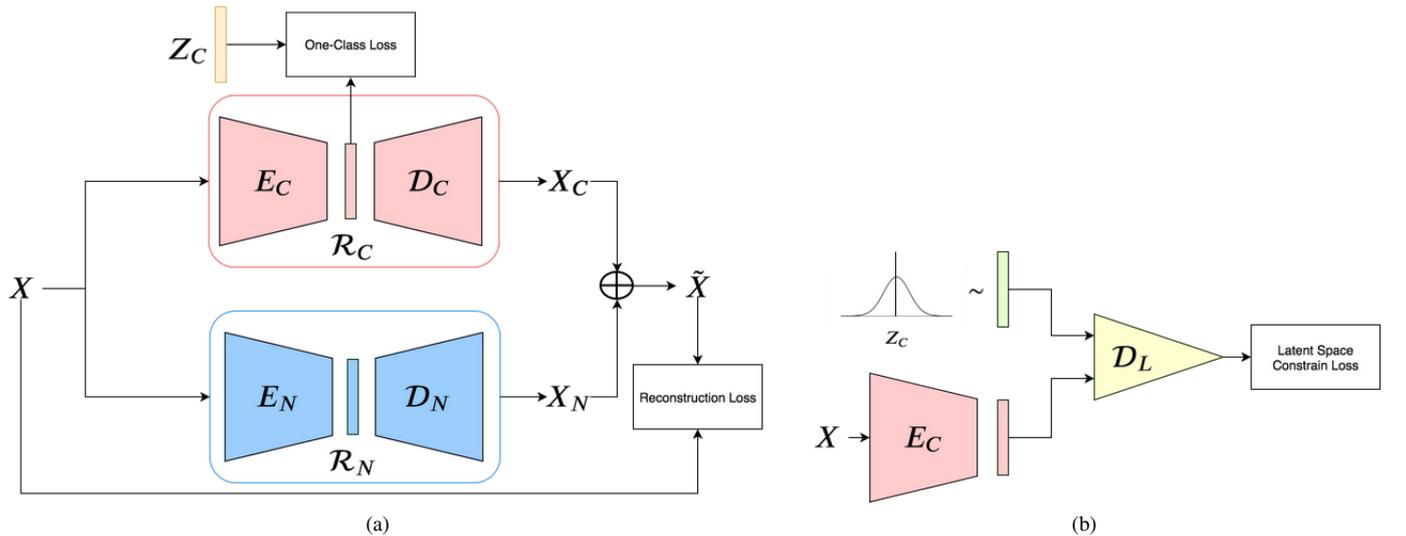

   \subfloat[][]{
      \includegraphics[scale=0.4]{DualReconstruction.png}
      \centering
      \label{fig:overall_a}
   }
   \subfloat[][]{
      \includegraphics[scale=0.4]{Aux.png}
      \centering
      \label{fig:overall_b}
   }
   \caption{The illustrations of three networks of the proposed method: (a) the reconstruction networks $\mathcal{R}_C$ and $\mathcal{R}_N$; (b) the auxiliary discriminator network $\mathcal{D}_L$.}%
   \label{fig:overall}%
\end{figure}



In the training phase, assuming the batch size is 1, a training image $X$ going through $\mathcal{R}_C$ and $\mathcal{R}_N$ is decomposed into its class-specific component $X_{C}$ and the non-class-specific component $X_{N}$, respectively. Motivated from deep SVDD \cite{ruff2018deep}, to make network $\mathcal{R}_C$ extract the common factors of variation of normal class training images, $\mathcal{R}_C$ is first trained to make the latent representations of all the training images cluster around a centroid $Z_C$ in the latent vector space. On the other hand, $\mathcal{R}_N$ is trained to construct a residual component to make the pixel-wise summation of $X_{C}$ and $X_{N}$ approximate the original input image $X$, i.e., minimizing the Euclidean distance between the $\tilde{X}$ and $X$, where $\tilde{X} = X_{C} + X_{N}$. We exploit a one-class loss and a reconstruction loss for the training of both $\mathcal{R}_C$ and $\mathcal{R}_N$, respectively as illustrated in Fig.~\ref{fig:overall_a}. Furthermore, to alleviate overfitting due to limited training images, we design an auxiliary network called latent space discriminator $\mathcal{D}_L$ which is trained together with the encoder of $\mathcal{R}_C$ in an adversarial learning strategy similar to GAN \cite{goodfellow2014generative}. Such auxiliary discriminator network is proposed to constrain the latent space distribution of normal class images to approximately satisfy a multivariate Gaussian distribution with its mean as the centroid $Z_C$ in the latent space. To this end, we propose a latent space constrained loss to train both $\mathcal{R}_C$ and $\mathcal{D}_L$ as shown in Fig.~\ref{fig:overall_b}.

 During inference, a testing image can be decomposed into its normal class component and the non-normal class component by simply going through $\mathcal{R}_C$ and $\mathcal{R}_N$, respectively. Then the anomaly score is calculated only based on the normal class component. Thereby, the anomaly detection can be performed via thresholding the anomaly score. Since our method has two reconstruction networks for class-specific image decomposition, it is named as dual deep reconstruction network based image decomposition, $\text{DDR-ID}$ for short. The details of the network architectures, the training/testing procedures and the anomaly score definitions are described in the following subsections.


\subsection{The architecture of $\mathcal{R}_C$ ,$\mathcal{R}_N$ and $\mathcal{D}_L$}

$\mathcal{R}_C$ and $\mathcal{R}_N$ exploit the same encodoer-decoder network architecture based on Convolution Neural Network ($\text{CNN}$). Inspired by the recent advances in representation learning with deep generative adversarial networks ($\text{DCGAN}$) \cite{radford2015unsupervised}, we specify the reconstruction networks using $\text{DCGAN}$-like architecture. The encoder consists of multiple convolution layers and then a fully-connected layer to embed an image into a lower dimensional latent vector. Each convolution layer is followed by a leakyReLu activation function and a batch normalization \cite{ioffe2015batch} layer. The decoder is constructed by multiple fractionally-strided convolution (deconvolution) layers \cite{zeiler2010deconvolutional} to map the latent vector to the reconstructed image. The deconvolution layers are also followed by leakyReLu activation functions and batch normalization layers. In general, the decoder structure can be treated as a ``transposed'' version of the encoder. Empirically, we observe that applying such $\text{GAN}$-like structures outperforms LeNet-type $\text{CNN}$ encoder-decoder structures adopted in~\cite{ruff2018deep}. Therefore, in practice, the architecture of our reconstruction networks can be replaced by any generator network of some GAN model as the decoder and construct the encoder by ``transposing'' the decoder, i.e., replacing the deconvolution layers by the convolution layers. The auxiliary network $\mathcal{D}_L$ exploits deep fully-connected networks. It consists of several blocks each consisting of a fully-connected layer, a ReLU activation function and a batch normalization layer. The final block of $\mathcal{D}_L$ uses the sigmoid activation function without batch normalization. The details of these network architectures are specified in Section \ref{Settings}.



\subsection{Training of DDR-ID}
Even though $\mathcal{R}_C$ and $\mathcal{R}_N$ are constructed using the same encoder-decoder architecture, the goals of training $\mathcal{R}_C$ and $\mathcal{R}_N$ are different. $\mathcal{R}_C$ is formulated to extract the normal class component while $\mathcal{R}_N$ is to extract the residual component. To this end, the training procedure consists of two stages: a symmetrical pretraining stage followed by an asymmetrical finetuning stage. The pretraining stage only trains $\mathcal{R}_C$ and $\mathcal{R}_N$ while the finetuning stage jointly
trains $\mathcal{R}_C$, $\mathcal{R}_N$ and the auxiliary $\mathcal{D}_L$.



\subsubsection{Pretraining stage}
In the pretraining stage, both $\mathcal{R}_C$ and $\mathcal{R}_N$ are trained as standard autoencoders for image reconstruction. Pretraining as an autoencoder to initialize the parameters of the networks serves for two goals: (i) to learn semantically meaningful and discriminative representations in both latent and reconstruction image space~\cite{xie2016unsupervised}; (ii) to select a good template (centroid) in the latent space. Specifically, considering we have $I$ training images from the normal class denoted as $\{X_1, X_2, \ldots ,X_I\}$, both $\mathcal{R}_C$ and $\mathcal{R}_N$ are pretrained by first solving the optimization problem of minimizing the reconstruction loss:
\begin{alignat}{2}
 \min _{ \theta _{ \mathcal{R}_C } }\frac{1}{I}{\sum _{ i }{ ||X_{ i }-\mathcal{R}_C } (X_{ i }; \theta _{ \mathcal{R}_C })||^{ 2 }},\\ \min _{ \theta _{ \mathcal{R}_N } }\frac{1}{I}{\sum _{ i }{ ||X_{ i }-\mathcal{R}_N } (X_{ i };\theta _{ \mathcal{R}_N })||^{ 2 }},
\end{alignat}
where $X_i$ denotes the $i_{th}$ training image, $\theta _{\mathcal{R}_C}$ and $\theta _{ \mathcal{R}_N }$ denote the parameters of network $\mathcal{R}_C$ and $\mathcal{R}_N$, respectively. $\mathcal{R}_C(X_i;\theta _{ \mathcal{R}_C })$ calculates the reconstruction image by feeding the image $X_i$ through the network $\mathcal{R}_C$. Since $\mathcal{R}_C$ is constructed by its encoder ${E}_C$ cascaded by the decoder $\mathcal{D}_C$, it is clear that $\mathcal{R}_C(X_i;\theta _{ \mathcal{R}_C }) = \mathcal{D}_C(Z_i;\theta _{ {D}_C })$ with $Z_i = {E}_C(X_i;\theta _{ {E}_C })$ representing the latent vector of $X_i$ encoded by $E_{C}$. Similarly, $\mathcal{R}_N(X_i;\theta _{ \mathcal{R}_N }) = {D}_N(Z^{'}_i;\theta _{ {D}_N })$ with $Z^{'}_i = {E}_N(X_i;\theta _{ {E}_N })$.

After pretraining, we calculate the mean $Z_C$ of all the latent vectors through encoding all the training images by ${E}_C$, i.e., $Z_C = \frac{1}{I}\sum_{i}^I{Z_i}$. $Z_C$ represents the normal class ``template'' in the latent space. The network parameters $\theta _{ \mathcal{R}_C}$ and $\theta _{ \mathcal{R}_N}$ are initialized after pretraining for the subsequent finetuning stage. This pretraining process is the same as Deep SVDD~\cite{ruff2018deep} except that we train two autoencoder networks simultaneously. We calculate the mean latent vector $Z_C$ only in the latent space of $\mathcal{R}_C$.

\subsubsection{Finetuning stage}
In the finetune stage, we further train the two reconstruction networks ($\mathcal{R}_C$ and $\mathcal{R}_N$) by minimizing three losses, namely the one-class loss, the auxiliary latent space constrain loss and the reconstruction loss. The one-class loss is the same as the second objective proposed by One-Class Deep SVDD \cite{ruff2018deep} which penalizes the Euclidean distance of each encoding latent representations of class-specific reconstruction network $\mathcal{R}_C$ to the mean latent vector $Z_C$ calculated in the pretraining stage:
\begin{equation}
\mathcal{L}_{OC}(E_{C})=\frac{1}{I}\sum_i{ ||E_{ C }(X_{ i };\theta _{ E_{ C } })-Z_{ C }||^{ 2 }}.
\end{equation}
By minimizing the one-class loss, the encoder $E_{C}$ is tuned to extract the common factors from the data variations related to the concept of the normal class. Following the Proposition 1 in \cite{ruff2018deep}, to avoid the encoder $E_{C}$ learning a constant function mapping, only the parameters of $E_{C}$ are tuned while the mean latent vector $Z_C$ preserves unchanged during finetuning. In addition, to alleviate overfitting of limited training images in the latent vector space, we train the encoder ${E}_C$ of $\mathcal{R}_C$ together with an auxiliary latent space discriminator $\mathcal{D}_L$ by minimizing a latent space constrained loss in an adversarial learning fashion similar to GAN \cite{goodfellow2014generative}. Specifically, given a normal class training sample $X$, we aim to constrain the latent representations encoded by ${E}_C$ to approximately follow a multivariate Gaussian distribution with its mean being set as the mean latent vector $Z_C$ and its standard deviation being set as a small value, e.g., $\mathcal{N}(Z_C, \sigma^2\bm{I})$. To this end, we formulate the following GAN-like latent space constrain loss:

\begin{equation}
\label{eq:minimaxgame-definition}
\mathcal{L}_{LSC}(E_{C}, \mathcal{D}_L) = \mathbb{E}_{Z \sim \mathcal{N}(Z_C, \sigma^2\bm{I})}[\log \mathcal{D}_L(Z; \theta _{\mathcal{D}_{L}})] + \mathbb{E}_{X \sim p_{\text{data}}}[\log (1 - \mathcal{D}_L(E_{C}(X; \theta _{{E}_{C}}); \theta _{\mathcal{D}_{L}}))],
\end{equation}
where $p_{\text{data}}$ refers to normal class data distribution. The minimax game with respect to the latent space constrain loss is formulated as $\min _{ \theta _{ E_C }}\max _{\theta _{\mathcal{D}_{L}}}{\mathcal{L}_{LSC}}$. The latent space constrain loss minimization is intuitively similar to the Deep SVDD with soft-bound objective as we train the class-specific encoder to generate latent vectors following a (small standard deviation) Gaussian distribution centered at $Z_C$.

Thirdly, we minimize the reconstruction loss to make the pixel-wise addition of the reconstruction images from $\mathcal{R}_C$ and $\mathcal{R}_N$ approximate the original input image:
\begin{equation}
\mathcal{L}_{R}(\mathcal{R}_C, \mathcal{R}_N)=\frac{1}{I}\sum_i{ ||\tilde{X_i} - X_i||^{ 2 }},
\end{equation}
where $\tilde{X_i} = \mathcal{R}_C(X_i;\theta _{ \mathcal{R}_C }) + \mathcal{R}_N(X_i;\theta _{ \mathcal{R}_N })$.

Overall, the parameters of $\mathcal{R}_C$ and $\mathcal{R}_N$ are finetuned to jointly optimize for the three losses by solving the following optimization problem:
\begin{equation}
\min _{ \theta _{ \mathcal{R}_C },\theta _{ \mathcal{R}_N } }\max _{\theta _{\mathcal{D}_{L}}}{\mathcal{L} _{ OC }+ \mathcal{L}_{LSC} +\mathcal{L} _{ R }}.
\end{equation}
{\color{black}Note that we use a simple summation of these three losses rather than a weighted summation as we empirically observe that a simple summation produces descent and stable performance. Hence, we do not introduce the hyperparamters to weight different losses.} Algorithm 1 summarizes the procedure for both pretraining and finetuning of the proposed DDR-ID.

\begin{algorithm}[ht]
\caption{Training of DDR-ID}
\begin{algorithmic}
\label{alg:AGF}
\STATE{\textbf{Input}: The normal class training samples.}
\STATE{\textbf{Pretraining}}
\STATE{Split the entire training set of normal samples into a training subset $\{ {X}_{1}, \dots, X_{I} \}$ and a validation subset $\{ {X}^{'}_{1}, \dots, X^{'}_{K} \}$.}
\FOR{number of pretraining iterations}
    \STATE{$\bullet$ Sample minibatch of $m$ samples $\{ {X}_{1}, \dots, X_{m} \}$ from the training subset.}
    \STATE{$\bullet$ Update $\mathcal{R}_C$ and $\mathcal{R}_N$ by descending the gradients of:
    \[
            \nabla_{\theta_{\mathcal{R}_C}} \frac{1}{m} \sum_{i=1}^m \left(
            { ||X_{ i }-\mathcal{R}_C } (X_{ i }; \theta _{ \mathcal{R}_C })||^{ 2 }
            \right),
    \]
    \[
            \nabla_{\theta_{\mathcal{R}_N}} \frac{1}{m} \sum_{i=1}^m \left(
            { ||X_{ i }-\mathcal{R}_N } (X_{ i }; \theta _{ \mathcal{R}_N })||^{ 2 }
            \right).
    \]}
    \ENDFOR
    \STATE{Calculate the normal-class-specific mean latent vector $Z_C$ by:
    \[
        Z_C = \frac{1}{I}\sum_{i}^I{Z_i}
    \]}
    \STATE{\textbf{Finetuning}}
    \FOR{number of finetuning iterations}
      \STATE{$\bullet$ Sample minibatch of $m$ latent samples $\{ Z_{1}, \dots, Z_{m} \}$ from $\mathcal{N}(Z_C, \sigma^2\bm{I})$}
      \STATE{$\bullet$ Sample minibatch of $m$ samples $\{ {X}_{1}, \dots, X_{m} \}$ from the training subset.}
      \STATE{$\bullet$ Calculate $m$ reconstruction samples $\{ \tilde{{X}_{1}}, \dots, \tilde{X_{m}} \}$ by $\tilde{X_i} = \mathcal{R}_C(X_i;\theta _{ \mathcal{R}_C }) + \mathcal{R}_N(X_i;\theta _{ \mathcal{R}_N })$.}
      \STATE{$\bullet$ Update $\mathcal{D}_L$ by ascending the gradient related to $\mathcal{L} _{LSC}$:
      \[
        \nabla_{\theta_{\mathcal{D}_L}} \frac{1}{m} \sum_{i=1}^m \left[
              \log \mathcal{D}_L(Z_i; \theta _{\mathcal{D}_{L}}) + \log (1 - \mathcal{D}_L(E_{C}(X_i; \theta _{{E}_{C}}); \theta _{\mathcal{D}_{L}}))
              \right].
      \]
      }
      \STATE{$\bullet$ Update $\mathcal{R}_C$ and $\mathcal{R}_N$ by descending the gradients related to $\mathcal{L} _{OC}$, $\mathcal{L} _{LSC}$ and $\mathcal{L} _{R}$:
      \[
        \nabla_{\theta_{\mathcal{R}_C}} \frac{1}{m} \sum_{i=1}^m \left[
              ||E_{ c }(X_{ i };\theta _{ E_{ C } })-Z_{ C }||^{ 2 }  + \log (1 - \mathcal{D}_L(E_{C}(X_i; \theta _{{E}_{C}}); \theta _{\mathcal{D}_{L}})) + ||\tilde{X_i} - X_i||^{ 2 }
              \right].
      \]
      \[
        \nabla_{\theta_{\mathcal{R}_N}} \frac{1}{m} \sum_{i=1}^m ||\tilde{X_i} - X_i||^{ 2 }.
      \]
      }

    \ENDFOR
    \STATE{ Here
        $\theta_{\mathcal{R}_C} = \{\theta _{ E_{ C }}, \theta _{ D_{ C }}\}$ and $\theta_{\mathcal{R}_N} = \{\theta _{ E_{ N }}, \theta _{ D_{ N }}\}.$}

\end{algorithmic}
\end{algorithm}

\subsection{Testing of DDR-ID}

After pretraining and finetuning, the class-specific decomposition of an image can be performed by simply feeding the image through both reconstruction networks. As $\mathcal{R}_C$ (along with $Z_C$) extracts the normal-class-specific information, we apply an anomaly score based classification rule for anomaly detection where the anomaly score can be defined in either the latent representation space or the reconstruction image space generated by $\mathcal{R}_C$. Specifically, given a testing image $X_t$, we propose two anomaly scores by exploiting our class-specific reconstruction network $\mathcal{R}_C$ and the mean latent vector $Z_C$.

\textbf{The latent anomaly score.} Similar to One-Class Deep SVDD~\cite{ruff2018deep}, we calculate the Euclidean distance from the encoded latent vector of $X^{t}$ to the mean latent vector $Z_C$ as the anomaly score, i.e.,
\begin{equation}
  AS_l(X^{t}) = ||E_C(X^{t};\theta_{E_C}) - Z_C||^2.
\label{AS_l}
\end{equation}
This definition is based on the assumption that the mean latent vector $Z_C$ represents the ``perfect'' template for the normal class in the latent space. Hence, if the latent representation of $X^{t}$ is closer to $Z_C$, $X^{t}$ is more probable to be labeled as normal.


\textbf{The reconstruction anomaly score.} The latent anomaly score predicts the anomality of a testing sample in the latent space. Compared with Deep SVDD~\cite{ruff2018deep} which drops the decoder part after pretraining, our class-specific reconstruction network $\mathcal{R}_C$ keeps tuning the decoder part in the finetuning stage. Therefore, we calculate a reconstruction anomaly score by measuring the distance between the class-specific reconstruction image $X^{t}$ and the template in the reconstruction image space. This template is generated by feeding the mean latent vector $Z_C$ through the encoder $E_C$. Thereby we define another anomaly score as:

\begin{equation}
AS_r(X^{t}) = ||\mathcal{R}_C(X^{t};\theta_{R_C}) - \mathcal{D}_C(Z_C;\theta_{\mathcal{D}_C})||^2.
\label{AS_r}
\end{equation}


\noindent As two anomaly score definitions are proposed, we adopt the one which produces the lower mean anomaly score across the samples in a hold-out validation set during training.



During testing, we use an anomaly score ($AS$) based criteria for anomaly detection. Given a testing image $X^{t}$ with its anomaly score being denoted as $AS(X^t)$, the decision of whether the testing image is predicted as normal can be made by thresholding $AS(X^t)$:
\begin{equation}
\begin{cases} \text{Normal\ Class}\quad \quad \quad \quad AS(X^{ t })<\tau,  \\ \text{Anomalous\ Class}\quad\quad   \text{Otherwise}.  \end{cases}
\label{OC_rule}
\end{equation}
where $\tau$ is the predefined threshold. Algorithm 2 summarizes the procedure of testing of DDR-ID.

\begin{algorithm}[ht]
\caption{Testing of DDR-ID}
\begin{algorithmic}
\label{alg:AGF}
\STATE{\textbf{Input}: A testing sample $X^{ t }$ and the validation set $\{ {X}^{'}_{1}, \dots, X^{'}_{K} \}$ split out in the training phase.}
\STATE{Calculate the latent anomaly score for each of the validation sample as $\{ AS_l({X}^{'}_{1}), \dots, AS_l(X^{'}_{K}) \}$ using Eq. (\ref{AS_l}).}
\STATE{Calculate the reconstruction anomaly score for each of the validation sample as $\{ AS_r({X}^{'}_{1}), \dots, AS_r(X^{'}_{K}) \}$ using Eq. (\ref{AS_r}).}
\STATE{Select the anomaly score definition by:
\[
  AS(X) = \begin{cases} AS_l(X),\quad  \frac{1}{K} \sum_{k=1}^K AS_l({X}^{'}_{k}) < \frac{1}{K} \sum_{k=1}^K AS_r({X}^{'}_{k}).\\ AS_r(X)\quad\    \text{Otherwise}.  \end{cases}
\]
}
\STATE{Calculate the anomaly score $AS(X^t)$ for $X^{ t }$.}
\STATE{Apply the classification rule for anomaly detection by Eq. (\ref{OC_rule}).}

\end{algorithmic}
\end{algorithm}

\section{Experiments}
In this section, we evaluate and analyze the anomaly detection performance of our method on three tasks: (i) anomaly detection on MNIST~\cite{lecun2010mnist} and CIFAR-10~\cite{krizhevsky2009learning} datasets; (ii) anomaly detection on biomedical patterns using Endosome~\cite{lin2019two} dataset; (iii) detection of adversarial attacks on GTSRB~\cite{stallkamp2011german} stop sign datasets. We compare our method against several related AD benchmarking methods. Since anomaly detection of images can be treated as binary classification, i.e., predicting a testing image as the normal class or not, the performance of the methods can be evaluated using the receiver operating characteristic (ROC) curve and the area under the ROC curve (AUC)~\cite{bradley1997use}. {\color{black}{To establish a fair comparison, we adopt the AUC statistics for the benchmarking methods reported in the original publications~\cite{ruff2018deep,chalapathy2018anomaly,sabokrou2018adversarially,deecke2018image}. We obtain the ROC curves by re-implementing the related benchmarking methods using the same architecture and parameter settings specified in the original publications.}} Our implementations are based on Pytorch framework~\cite{paszke2017automatic} and Python3 using a NVIDIA TITAN X GPU.

\subsection{The Settings of Compared Methods}
\label{Settings}
In this subsection, we specify the architectures and the parameters for the benchmarking methods and our $\text{DDR-ID}$ in the experiments. For training, the batch size is set at $256$ and the weight-decay factor is set at $10^{-6}$. The Adam optimizer is used with the recommended parameters in \cite{kingma2014adam}. The specific settings for each method are listed as follows.
\begin{itemize}
    \item \textbf{Anomaly Detection with Generative Adversarial Networks ($\text{AnoGAN}$)~\cite{schlegl2017unsupervised}.} We report the performance of AnoGAN from \cite{ruff2018deep} where $\text{DCGAN}$ structure is applied as the backbone for $\text{AnoGAN}$ and the latent space dimensionality is set at $256$. The anomaly score is calculated as the overall loss, i.e., the summation of the residual loss and the discriminant loss~\cite{schlegl2017unsupervised}.

    \item \textbf{Adversarially Learned One-Class Classifier ($\text{ALOCC}$)~\cite{sabokrou2018adversarially}.} We implement $\text{ALOCC}$ network $\mathcal{R}$ based on the similar encoder-decoder structure as our own reconstruction networks. Regarding the discrimination network $D$, we exploit the discriminator of $\text{DCGAN}$. The number of training epochs is set at $50$ and the learning rate is set at $10^{-5}$. The anomaly score is defined as the reconstruction error of $\mathcal{R}$ network.


    \item \textbf{Deep Convolution Autoencoder ($\text{DCAE}$)}. We build LeNet-type DCAE exploited in \cite{ruff2018deep} as our benchmarking convolution autoencoder method. The $\text{DCAE}$ is trained by minimizing $\text{MSE}$ loss. The number of epochs is set at $150$. The anomaly score is defined as the reconstruction error.

    {\color{black}{
    \item \textbf{One-Class Neural Network ($\text{OC-NN}$)~\cite{chalapathy2018anomaly}}. For $\text{OC-NN}$, we reported the average AUC values from \cite{chalapathy2018anomaly} on MNIST and CIFAR-10 datasets.

    \item \textbf{Anomaly Detection with Generative Adversarial Networks ($\text{ADGAN}$)~\cite{deecke2018image}}. We compare our method with the average AUC of $\text{ADGAN}$ reported in \cite{deecke2018image}. As \cite{deecke2018image} conducted a different experimental protocol, we ran our method under the same protocol and then compare the AUC results with ones reported in \cite{deecke2018image}.
    }}

    \item \textbf{Deep Support Vector Data Description (Deep SVDD)~\cite{ruff2018deep}.} We implement two models, namely Soft-Bound Deep SVDD and One-Class Deep SVDD by exploiting the same LeNet-type CNNs for both models as described in~\cite{ruff2018deep}. For Soft-Bound SVDD, the parameter $\upsilon$ for the upper bound on the fraction of abnormal data is set at $0.05$ and the radius $R$ is solved via line search every $5$ epochs. The number of epochs for pretraining the autoencoder and finetuning are $150$ and $100$, respectively. The two-phase learning rate schedule is applied for both pretraining and finetuning with the initial learning rate at $10^{-4}$ and set at $10^{-5}$ after $50$ epochs. Similar to \cite{ruff2018deep}, all the images are preprocessed with global contrast normalization using the L1 norm and rescaled to $[0,1]$ via min-max-scaling. The anomaly score is calculated by the Euclidean distance between the encoded vector to the mean latent vector.


    \item \textbf{Our Dual Reconstruction Networks Based Image Decomposition ($\text{DDR-ID}$).} The network architectures for the networks in $\mathcal{R}_C$, $\mathcal{R}_N$ and $\mathcal{D}_L$ are specified in Table~\ref{tab:AE_D_Structure}. We use the same pretraining and finetuning parameters (training epochs and learning rate schedule) of Deep SVDD \cite{ruff2018deep} and adopt the same global contrast normalization for preprocessing. We split the entire training set into 90$\%$ samples in the training subset and the remaining 10$\%$ for validation subset for the selection of anomaly score.

\end{itemize}

{\color{black}{
\begin{table}[htbp]\scriptsize
  \centering
  \caption{The network architectures for the encoder, the decoder and the discriminator within $\mathcal{R}_C$, $\mathcal{R}_N$ and $\mathcal{D}_L$}
  \begin{threeparttable}
    \begin{tabular}{|c|c|c|rrrrrr|}
    \toprule
          & MNIST & CIFAR-10\textbackslash{}Endosome\textbackslash{}GTRSB & \multicolumn{1}{r|}{} & \multicolumn{2}{c|}{MNIST\textbackslash{}CIFAR-10\textbackslash{}Endosome\textbackslash{}GTRSB} & \multicolumn{1}{r|}{} & \multicolumn{1}{c|}{MNIST} & \multicolumn{1}{c|}{CIFAR-10\textbackslash{}Endosome\textbackslash{}GTRSB} \\
    \midrule
    \multirow{12}[6]{*}{Encoder} & \multicolumn{2}{c|}{Input Image} & \multicolumn{1}{c|}{\multirow{12}[6]{*}{Decoder}} & \multicolumn{2}{c|}{Input Latent Vector} & \multicolumn{1}{c|}{\multirow{12}[6]{*}{Discriminator}} & \multicolumn{2}{c|}{Input Latent Vector} \\
\cmidrule{2-3}\cmidrule{5-6}\cmidrule{8-9}          & \multicolumn{2}{c|}{Conv4-2-64} & \multicolumn{1}{c|}{} & \multicolumn{2}{c|}{DeConv4-1-512} & \multicolumn{1}{c|}{} & \multicolumn{1}{c|}{FC-64} & \multicolumn{1}{c|}{FC-128} \\
\cmidrule{8-9}          & \multicolumn{2}{c|}{LeakyReLU-0.2} & \multicolumn{1}{c|}{} & \multicolumn{2}{c|}{BatchNorm} & \multicolumn{1}{c|}{} & \multicolumn{2}{c|}{BatchNorm} \\
          & \multicolumn{2}{c|}{Conv4-2-128} & \multicolumn{1}{c|}{} & \multicolumn{2}{c|}{LeakyReLU-0.2} & \multicolumn{1}{c|}{} & \multicolumn{2}{c|}{ReLU} \\
          & \multicolumn{2}{c|}{BatchNorm} & \multicolumn{1}{c|}{} & \multicolumn{2}{c|}{DeConv4-2-256} & \multicolumn{1}{c|}{} & \multicolumn{2}{c|}{FC-32} \\
          & \multicolumn{2}{c|}{LeakyReLU-0.2} & \multicolumn{1}{c|}{} & \multicolumn{2}{c|}{BatchNorm} & \multicolumn{1}{c|}{} & \multicolumn{2}{c|}{BatchNorm} \\
          & \multicolumn{2}{c|}{Conv4-2-256} & \multicolumn{1}{c|}{} & \multicolumn{2}{c|}{LeakyReLU-0.2} & \multicolumn{1}{c|}{} & \multicolumn{2}{c|}{ReLU} \\
          & \multicolumn{2}{c|}{BatchNorm} & \multicolumn{1}{c|}{} & \multicolumn{2}{c|}{DeConv4-2-128} & \multicolumn{1}{c|}{} & \multicolumn{2}{c|}{FC-16} \\
          & \multicolumn{2}{c|}{LeakyReLU-0.2} & \multicolumn{1}{c|}{} & \multicolumn{2}{c|}{BatchNorm} & \multicolumn{1}{c|}{} & \multicolumn{2}{c|}{BatchNorm} \\
          & \multicolumn{2}{c|}{Conv4-2-512} & \multicolumn{1}{c|}{} & \multicolumn{2}{c|}{LeakyReLU-0.2} & \multicolumn{1}{c|}{} & \multicolumn{2}{c|}{ReLU} \\
          & \multicolumn{2}{c|}{BatchNorm} & \multicolumn{1}{c|}{} & \multicolumn{2}{c|}{DeConv4-2-64} & \multicolumn{1}{c|}{} & \multicolumn{2}{c|}{FC-1} \\
          & \multicolumn{2}{c|}{LeakyReLU-0.2} & \multicolumn{1}{c|}{} & \multicolumn{2}{c|}{Tanh Activation} & \multicolumn{1}{c|}{} & \multicolumn{2}{c|}{Sigmoid Activation} \\
    \midrule
    Latent Space & FC-128 & FC-512 &       &       &       &       &       &  \\
    \bottomrule
    \end{tabular}%
  \begin{tablenotes}
      \footnotesize
      \item \textbf{Conv} denotes convolution layer with the notation Conv$<$kernel dimension$>$-$<$kernel stride$>$-$<$number of output channels$>$.
      \item \textbf{BatchNorm} denotes the batch normalization operation.
      \item \textbf{FC} denotes the fully-connected layer with notation FC-$<$number of output channels$>$.
      \item \textbf{DeConv} denotes the deconvolution layer with the notation DeConv$<$kernel dimension$>$-$<$kernel stride$>$-$<$number of output channels$>$.
      \item \textbf{LeakyReLU} denotes the leaky ReLU activation function with notation LeakyReLU-$<$negative slope$>$.
      \item \textbf{ReLU} denotes the ReLU activation function.
    \end{tablenotes}
    \end{threeparttable}
    \label{tab:AE_D_Structure}%
\end{table}%
}}

\subsection{Anomaly Detection on MNIST and CIFAR-10 datasets}\label{MNISTandCIFAREx}

In this subsection , we conduct the AD experiments on $\text{MNIST}$ and $\text{CIFAR-10}$. Both $\text{MNIST}$ and $\text{CIFAR-10}$ consist of ten classes. For each experiment, one class is selected as the normal class and the others are treated as the anomalous classes. During training, only the normal class training images are used. The training set size is around $6000$ for $\text{MNIST}$ and $5000$ for $\text{CIFAR-10}$ while the testing set size is $10000$. For each normal class, we conduct 10 rounds of experiments to plot the average ROC curves and calculate the average AUCs across different normal classes. {\color{black}{In our experiments, we adopt two protocols. For the comparison with most benchmarking methods, we follow the protocol used in~\cite{ruff2018deep} where all the testing images are adopted for evaluation. For the comparison with ADGAN, since its performance was tested using randomly selected 5000 images from the testing set instead of the whole testing set~\cite{deecke2018image}. We tested our DDR-ID under this protocol and compare the average AUCs with the ones reported in~\cite{deecke2018image}}}.

\begin{table}[htpb]\small
  \centering
  \caption{Average AUCs in \% over 10 rounds of experiments under the protocol adopted in~\cite{ruff2018deep}. The best AUC value is in bold.}
    \begin{tabular}{|p{15.5em}|c|c|}
    \toprule
    \multicolumn{1}{|c|}{\multirow{2}[3]{*}{Method}} & \multicolumn{2}{c|}{Datasets} \\
\cmidrule{2-3}    \multicolumn{1}{|c|}{} & $\text{MNIST}$ & CIFAR-10 \\
\midrule
    \multicolumn{1}{|c|}{OC-SVM/SVDD~\cite{Tax2004}} & 91.3  & 64.8 \\
    \multicolumn{1}{|c|}{KDE~\cite{10.2307/2237880}} & 86.8  & 64.9 \\
    \multicolumn{1}{|c|}{IF~\cite{liu2008isolation}} & 92.6  & 55.5 \\
    \multicolumn{1}{|c|}{ALOCC~\cite{sabokrou2018adversarially}} & 84.2  & 54.3 \\
    \multicolumn{1}{|c|}{AnoGAN~\cite{schlegl2017unsupervised}} & 91.3  & 61.8 \\
    \multicolumn{1}{|c|}{DCAE~\cite{ruff2018deep}} & 91.4  & 57.5 \\
    \multicolumn{1}{|c|}{OC-NN~\cite{chalapathy2018anomaly}} & 92.3  & 61.1 \\
    \multicolumn{1}{|c|}{SOFT-BOUND DEEP SVDD~\cite{ruff2018deep}} & 93.5  & 63.3 \\
    \multicolumn{1}{|c|}{ONE-CLASS DEEP SVDD~\cite{ruff2018deep}} & 94.8  & 64.8 \\
    \multicolumn{1}{|c|}{Our DDR-ID} & \textbf{96.2}  & \textbf{65.4} \\
    \bottomrule
    \end{tabular}%
  \label{tab:AUCs}%
\end{table}%

\begin{table}[htpb]\small
  \centering
  \caption{Average AUCs in \% over 10 rounds of experiments under the protocol adopted in~\cite{deecke2018image}. The best AUC value is in bold.}
    \begin{tabular}{|p{15.5em}|c|c|}
    \toprule
    \multicolumn{1}{|c|}{\multirow{2}[3]{*}{Method}} & \multicolumn{2}{c|}{Datasets} \\
\cmidrule{2-3}    \multicolumn{1}{|c|}{} & MNIST & CIFAR-10 \\
\midrule
    \multicolumn{1}{|c|}{ADGAN[29]} & \textbf{96.8}  & 63.4 \\
    \multicolumn{1}{|c|}{Our DDR-ID} & 96.6  & \textbf{65.9} \\
    \bottomrule
    \end{tabular}%
  \label{tab:AUCs_new}%
\end{table}%



\begin{figure}[htpb]
   \subfloat[][]{
      \includegraphics[scale=0.55]{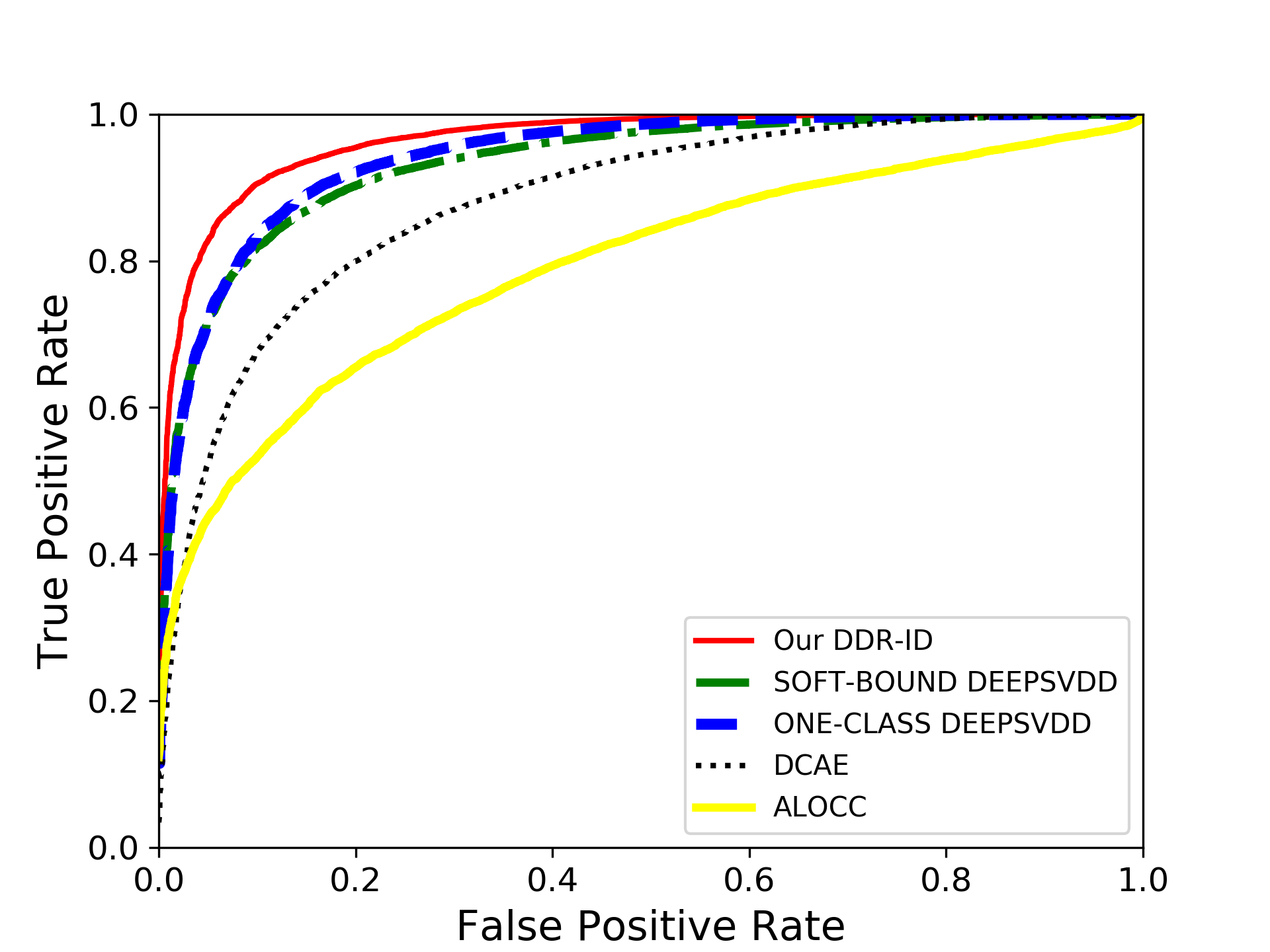}
      \centering
      \label{mnist_roc}
   }
   \subfloat[][]{
      \includegraphics[scale=0.55]{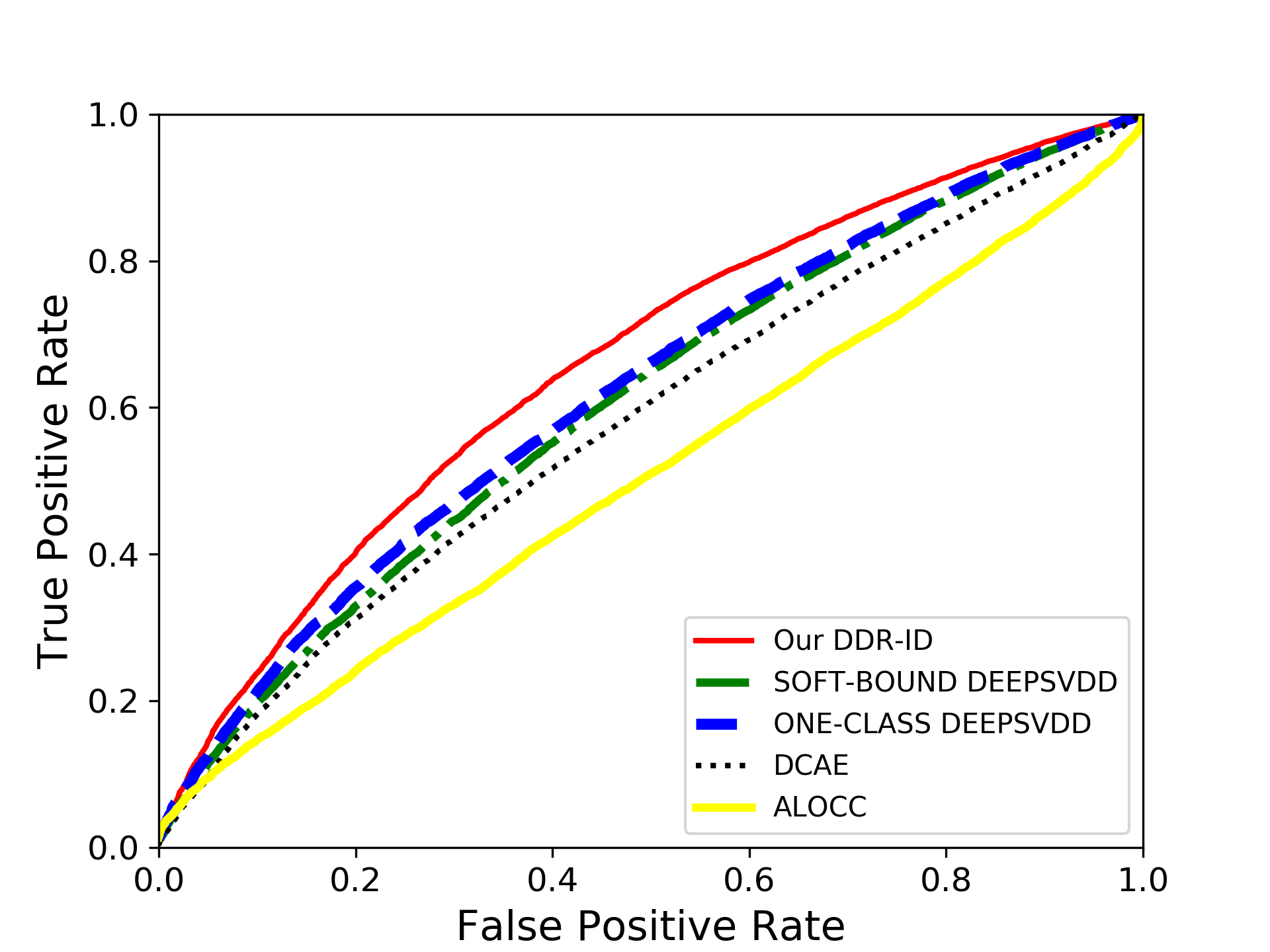}
      \centering
      \label{ciar_roc}
   }
   \caption{The average ROC curves plotted for deep convolution network related methods tested on (a) MNIST; (b) CIFAR-10. Best viewed in the color version.}%
   \label{ROC_Curves}%
\end{figure}
\begin{figure}[!t]
   \subfloat[][]{
      \includegraphics[scale=0.28]{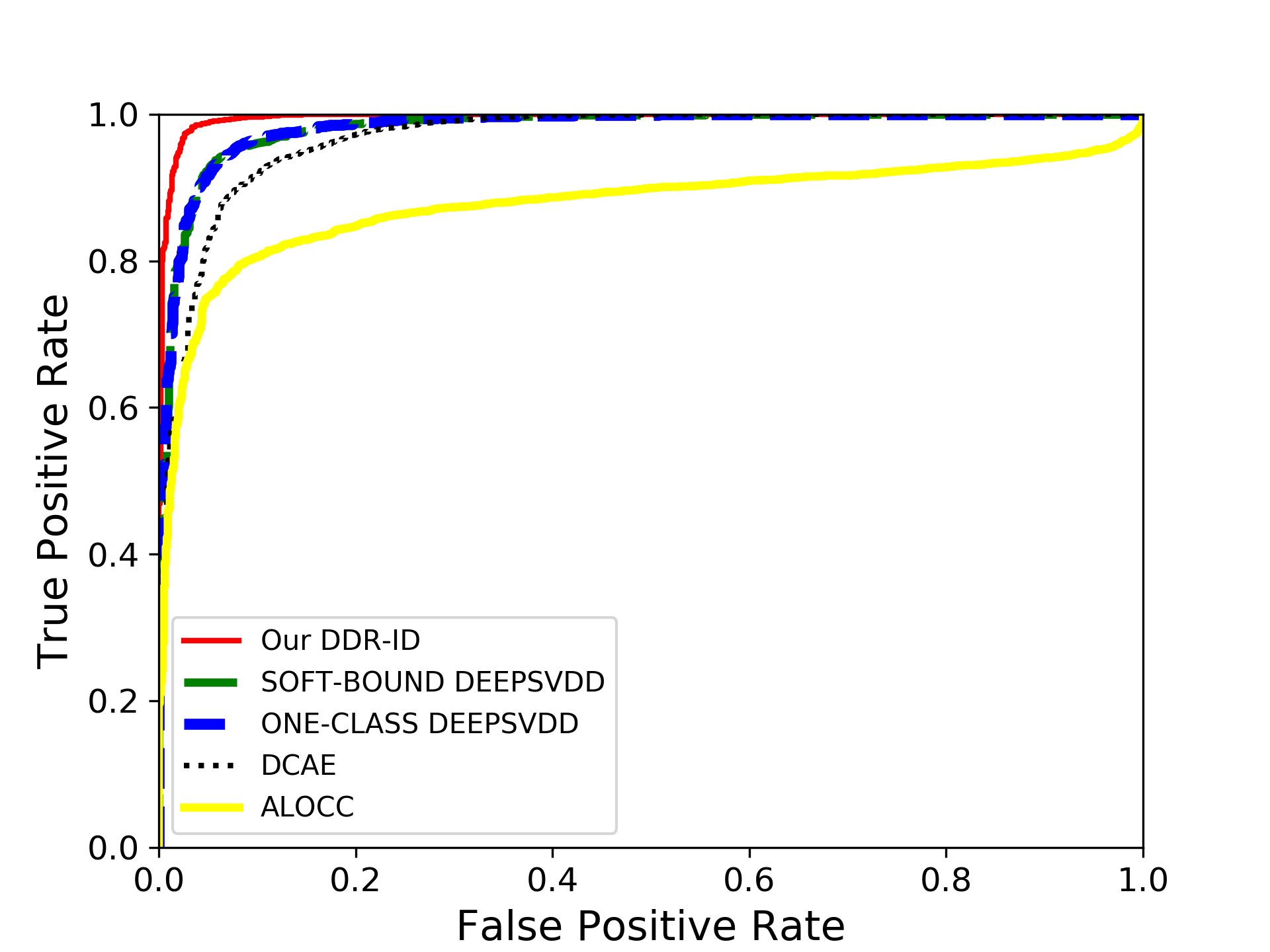}
      \centering
      \label{mnist_0}
   }
   \subfloat[][]{
      \includegraphics[scale=0.28]{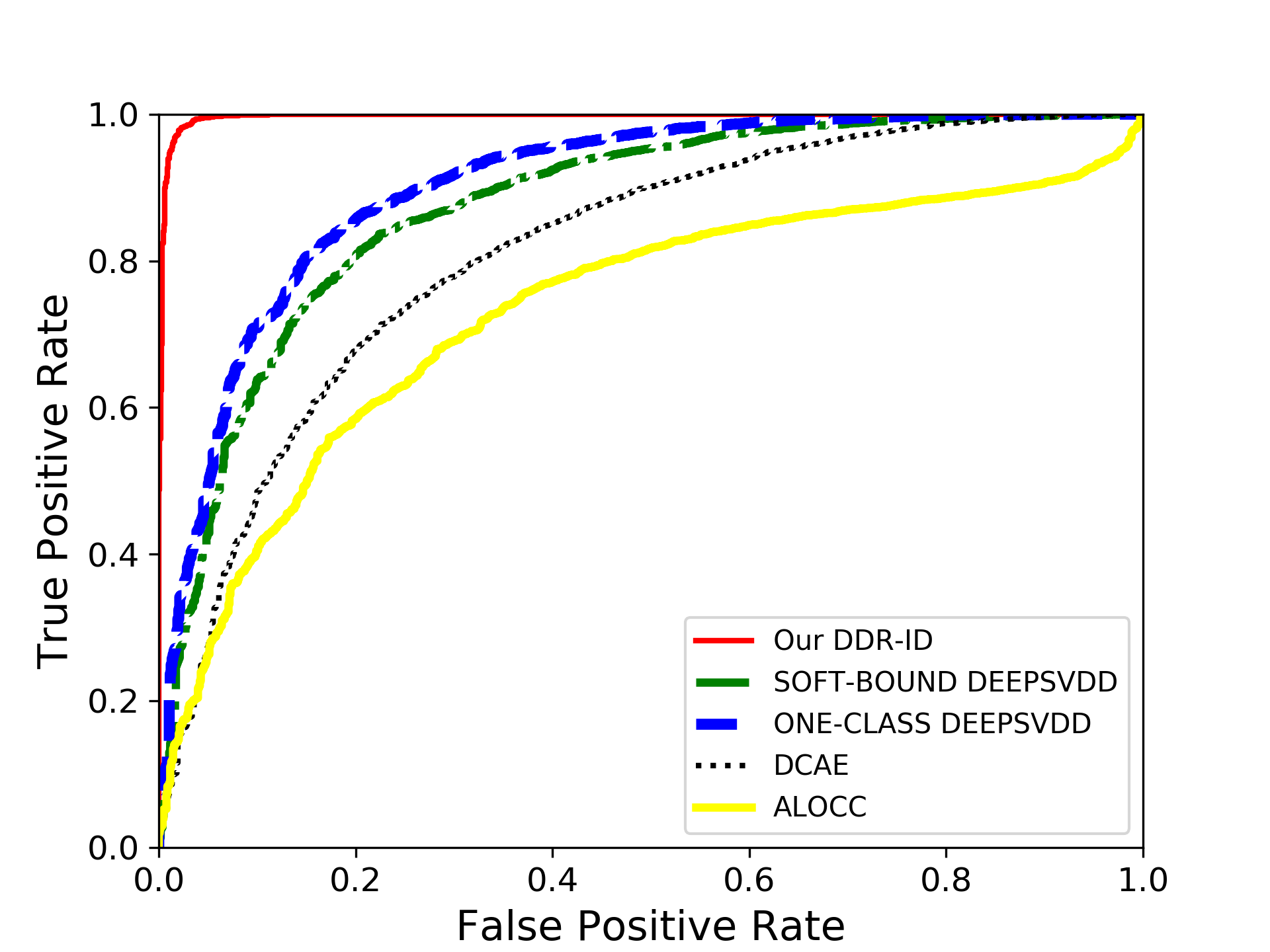}
      \centering
      \label{mnist_2}
   }
   \subfloat[][]{
      \includegraphics[scale=0.28]{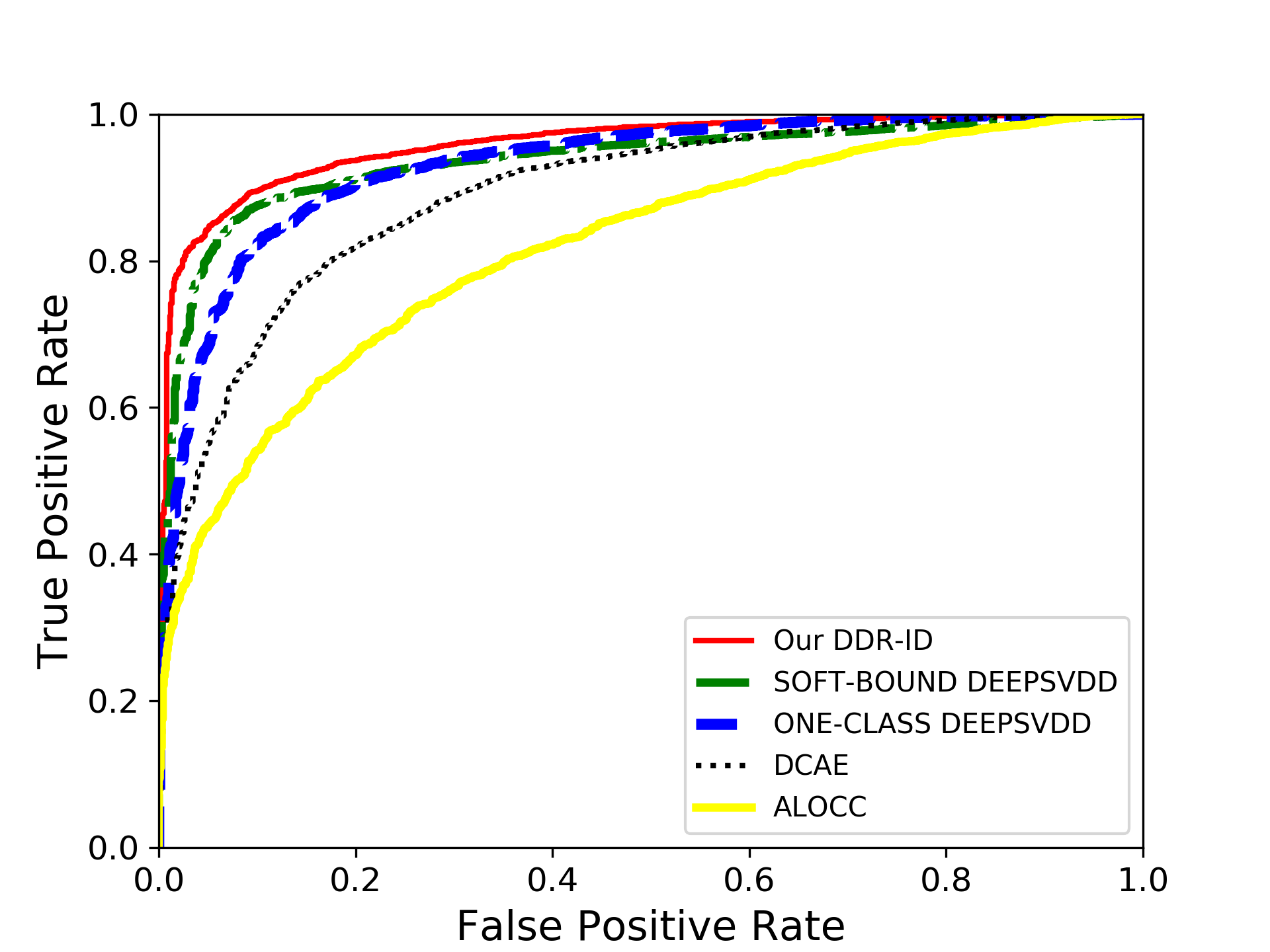}
      \centering
      \label{mnist_4}
   }
   \subfloat[][]{
      \includegraphics[scale=0.28]{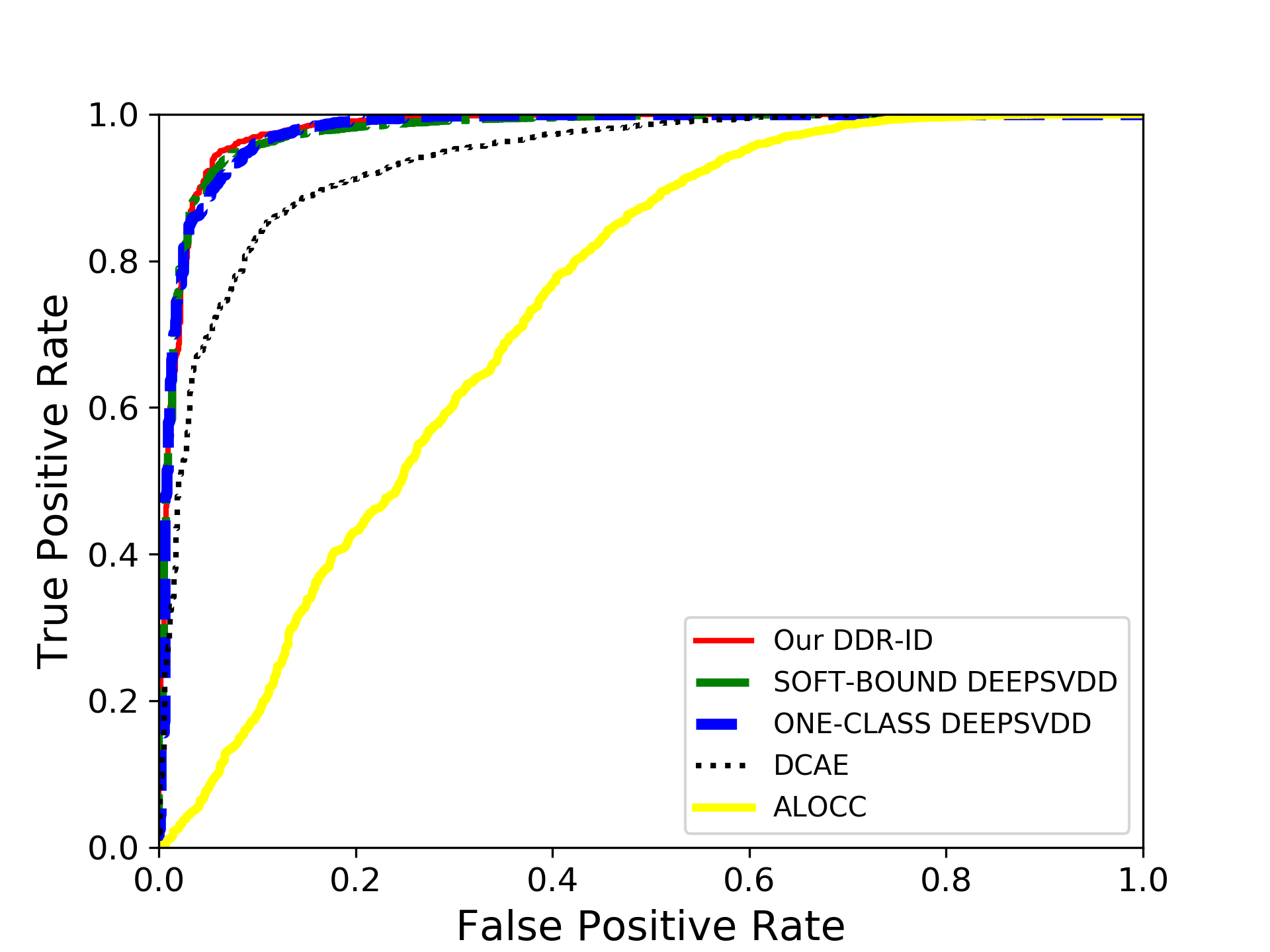}
      \centering
      \label{mnist_6}
   }\\

   \subfloat[][]{
      \includegraphics[scale=0.28]{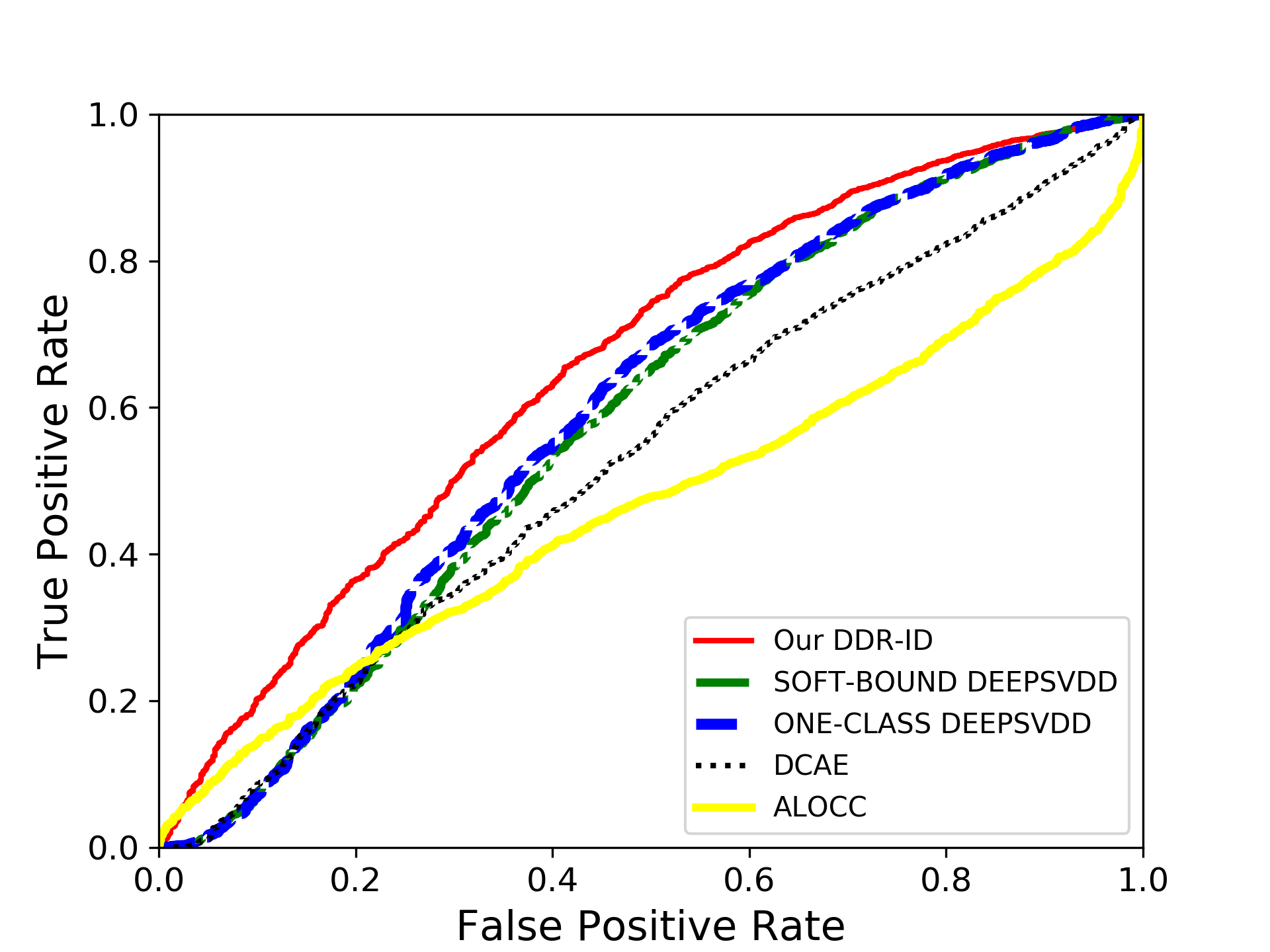}
      \centering
      \label{cifar_plane}
   }
   \subfloat[][]{
      \includegraphics[scale=0.28]{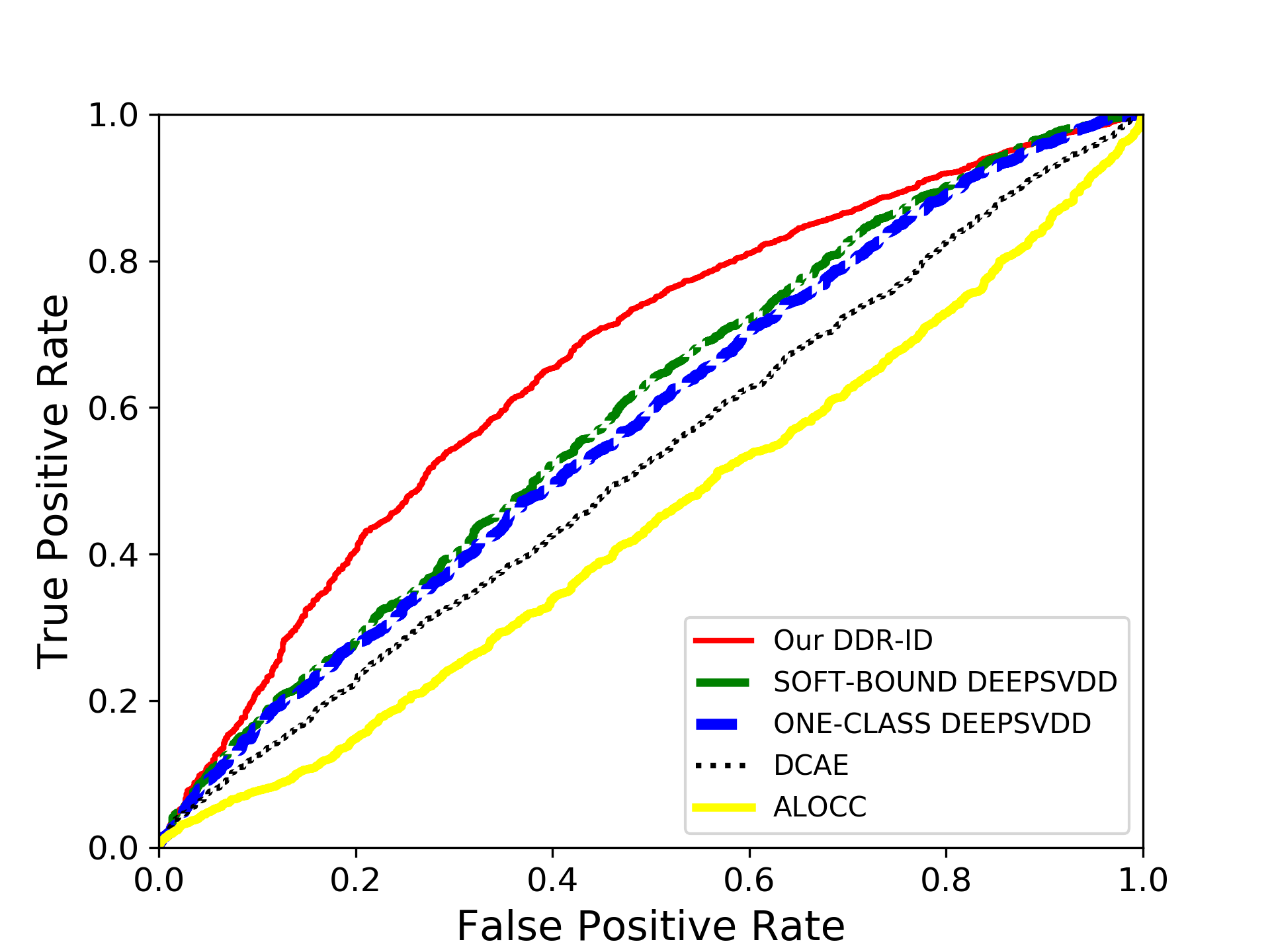}
      \centering
      \label{cifar_deer}
   }
   \subfloat[][]{
      \includegraphics[scale=0.28]{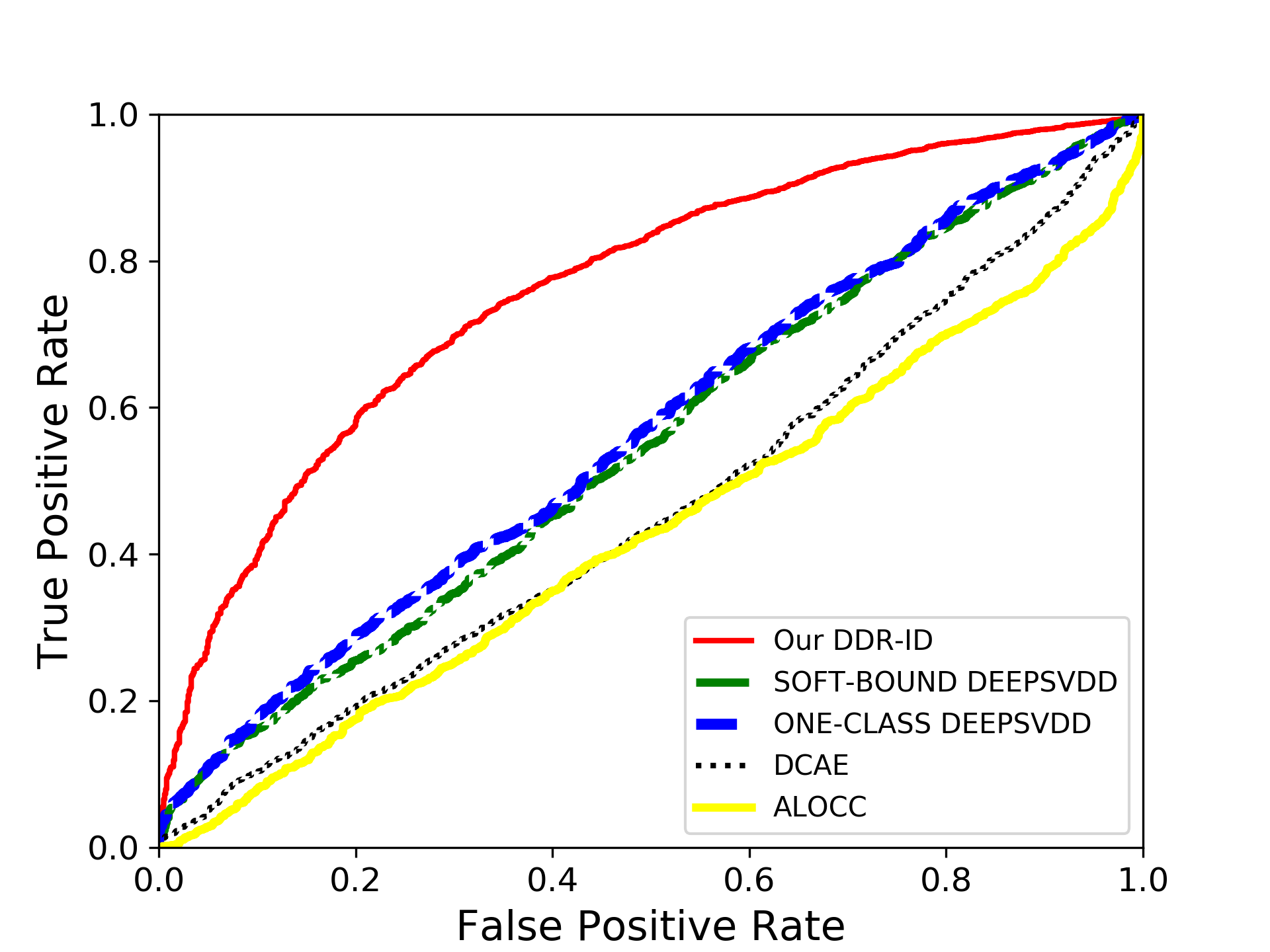}
      \centering
      \label{cifar_frog}
   }
   \subfloat[][]{
      \includegraphics[scale=0.28]{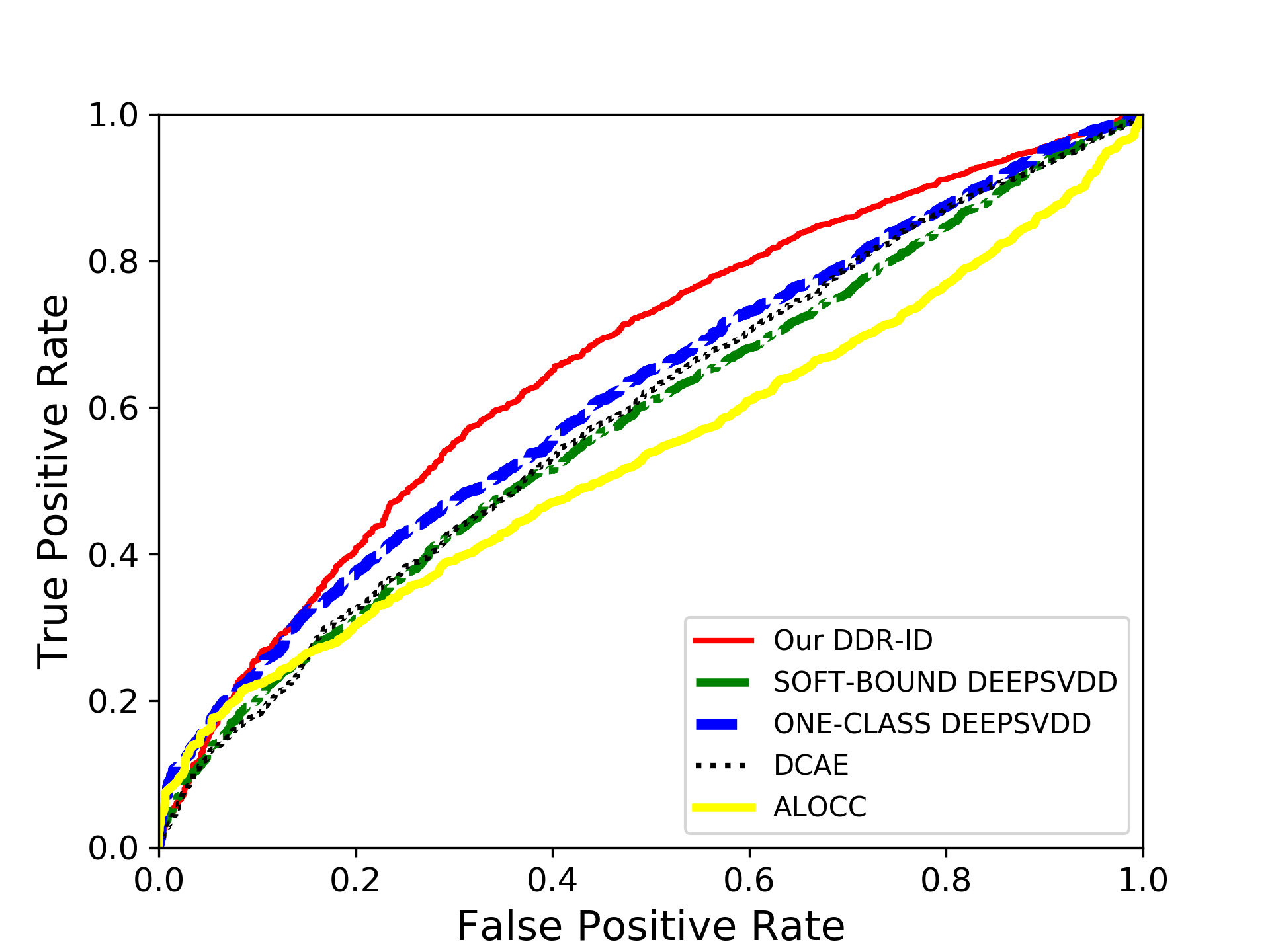}
      \centering
      \label{cifar_horse}
   }
   \caption{The ROC curves plotted for target class in MNIST selected as (a) ``0''; (b) ``2''; (c) ``4''; (d) ``6''; and in CIFAR-10 selected as (e) ``plane''; (f) ``deer''; (g) ``frog''; (h) ``horse''. Best viewed in the color version.}%
   \label{ROC_Curves_class}%
\end{figure}

{\color{black}{Table~\ref{tab:AUCs} presents the average AUCs over all the classes for both datasets under the protocol adopted in \cite{ruff2018deep}. The specific AUC values for each normal class are reported in the Section 1 of our supplementary material. Table~\ref{tab:AUCs_new} presents the average AUCs under the protocol adopted in ~\cite{deecke2018image}. Figs.~\ref{mnist_roc} and \ref{ciar_roc} illustrate the ROC curves for the deep convolution network related comparison methods based on our implementation with details specified in Section~\ref{Settings}. In addition, Fig.~\ref{mnist_0} to Fig.~\ref{cifar_horse} illustrate the ROC curves for different individual normal classes, namely digit ``0'', ``2'', ``4'' and ``6'' for MNIST and ``plane'', ``deer'', ``frog'' and ``horse'' for CIFAR-10. For the comparison under the protocol in ~\cite{ruff2018deep} (Table~\ref{tab:AUCs} and the ROC curves), it is observed that our DDR-ID generally outperforms all the benchmarking methods in terms of average AUCs and ROC performance. Specifically, the proposed DDR-ID outperforms three image reconstruction based methods (ALOCC, AnoGAN and DCAE) by exploiting only discriminative normal class component and filtering out non-normal-class component. Compared with Deep SVDD~\cite{ruff2018deep} which uses only the latent space to evaluate testing images' anomaly degree, the proposed DDR-ID can investigate the anomaly degree in both the normal class latent space and the reconstruction image space. For the comparison under the protocol in ~\cite{deecke2018image} (Table~\ref{tab:AUCs_new}), it is noted that the proposed DDR-ID produces comparable AUC in MNIST dataset (96.6\% vs 96.8\%) and better results in CIFAR-10 dataset (65.9\% vs 63.4\%) compared with ADGAN.}}

\begin{figure*}[!t]
   \subfloat[][]{
      \includegraphics[width = 3.5in,height=2in]{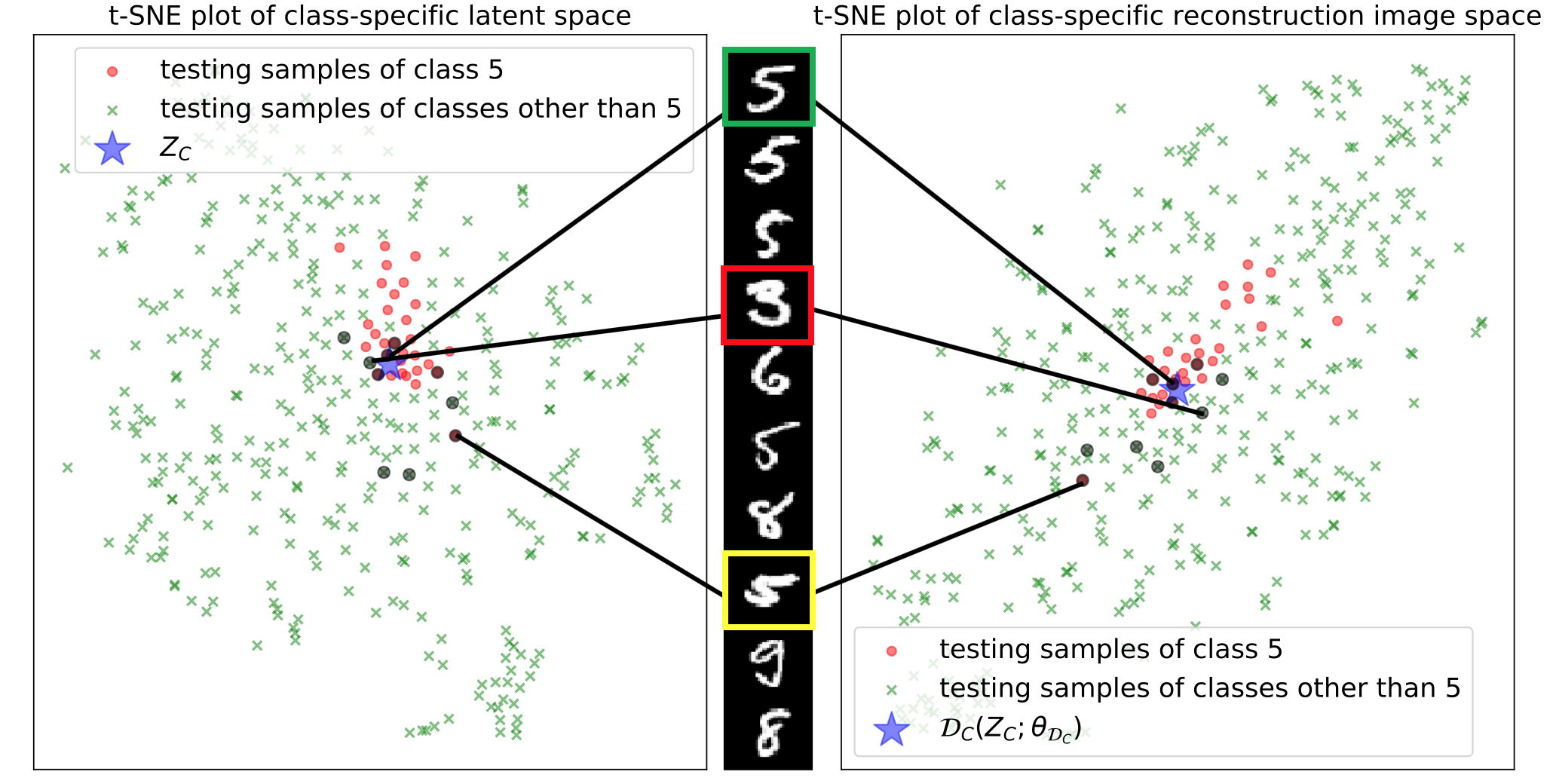}
      \centering
      \label{t-sne-5}
   }
   \subfloat[][]{
      \includegraphics[width = 3.5in,height=2in]{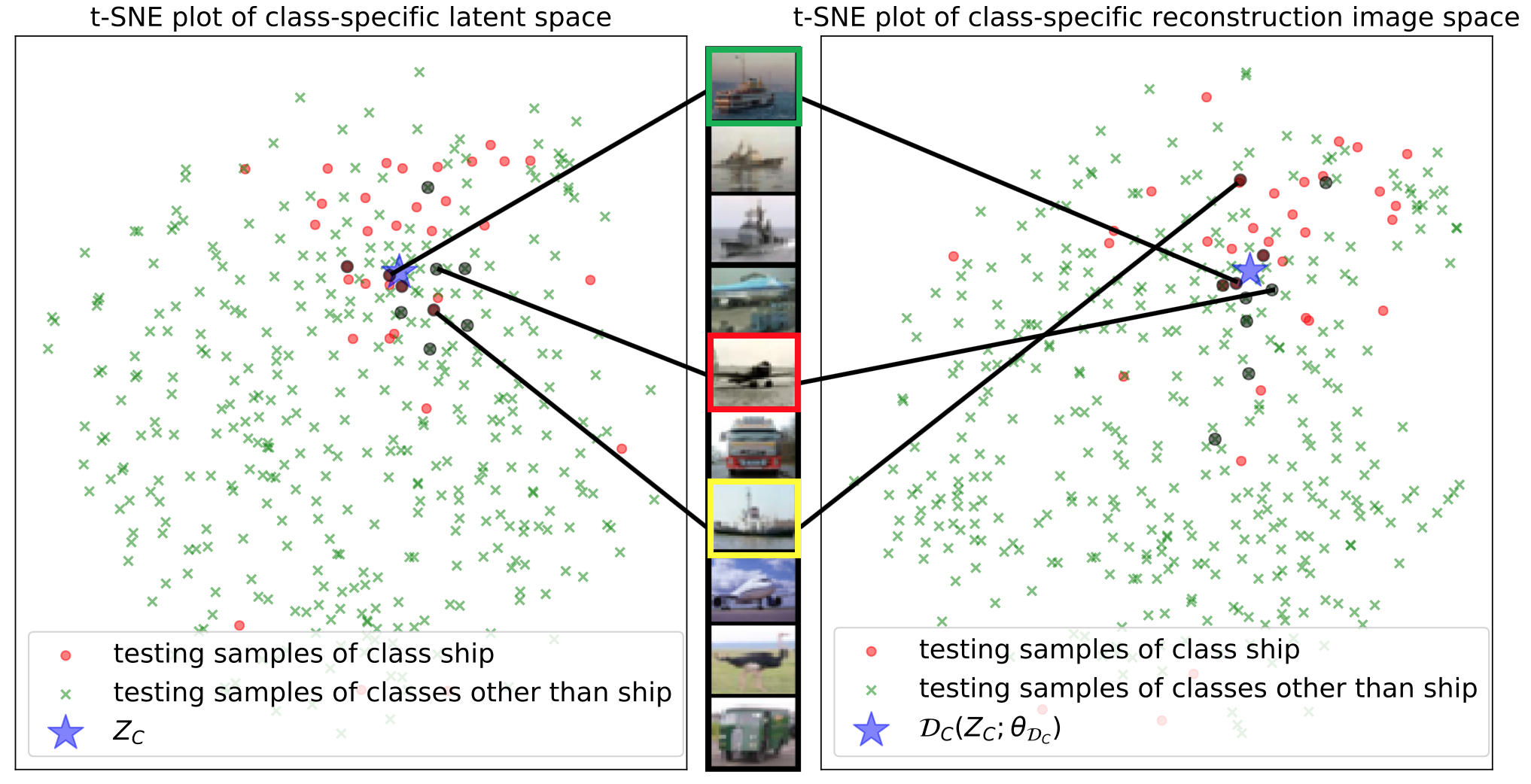}
      \centering
      \label{t-sne-ship}
   }
   \caption{2D t-SNE plots of class-specific latent representations (left) and class-specific reconstructed images (right) for 500 testing images. The example testing images (middle) are sorted in an ascending order with respect to anomaly score. The representations of the example images are shown in shadowed points. (a) Training using MNIST digit ``5'' as the normal class. (b) Training using CIFAR-10 class ``ship'' as the normal class. Best viewed in the color version.}%
   \label{t-sne-visualization}%
\end{figure*}
\begin{figure}[!t]
   \subfloat[][]{
      \includegraphics[scale=0.34]{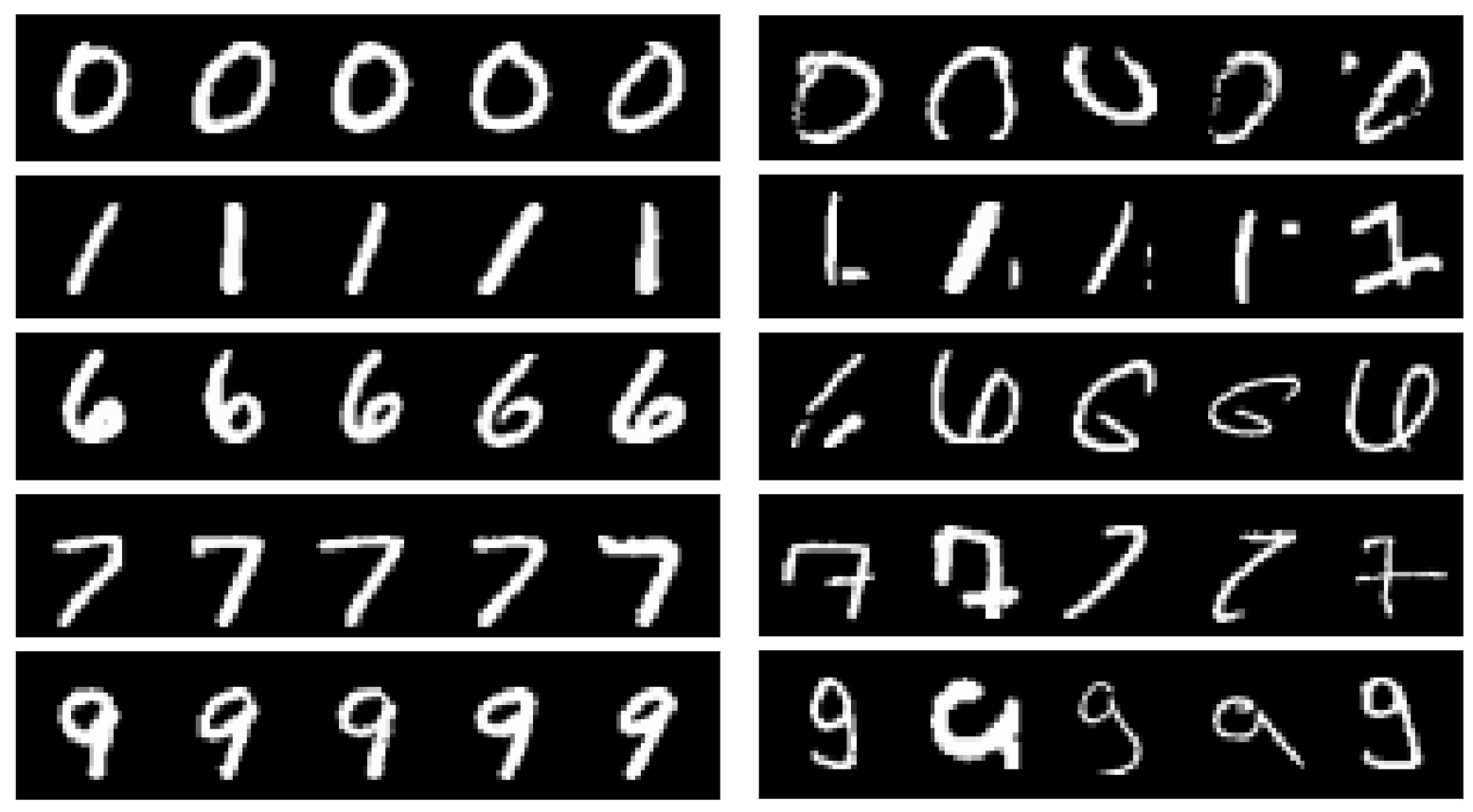}
      \centering
      \label{inlier_outlier_visualization_mnist}
   }
   \subfloat[][]{
      \includegraphics[scale=0.34]{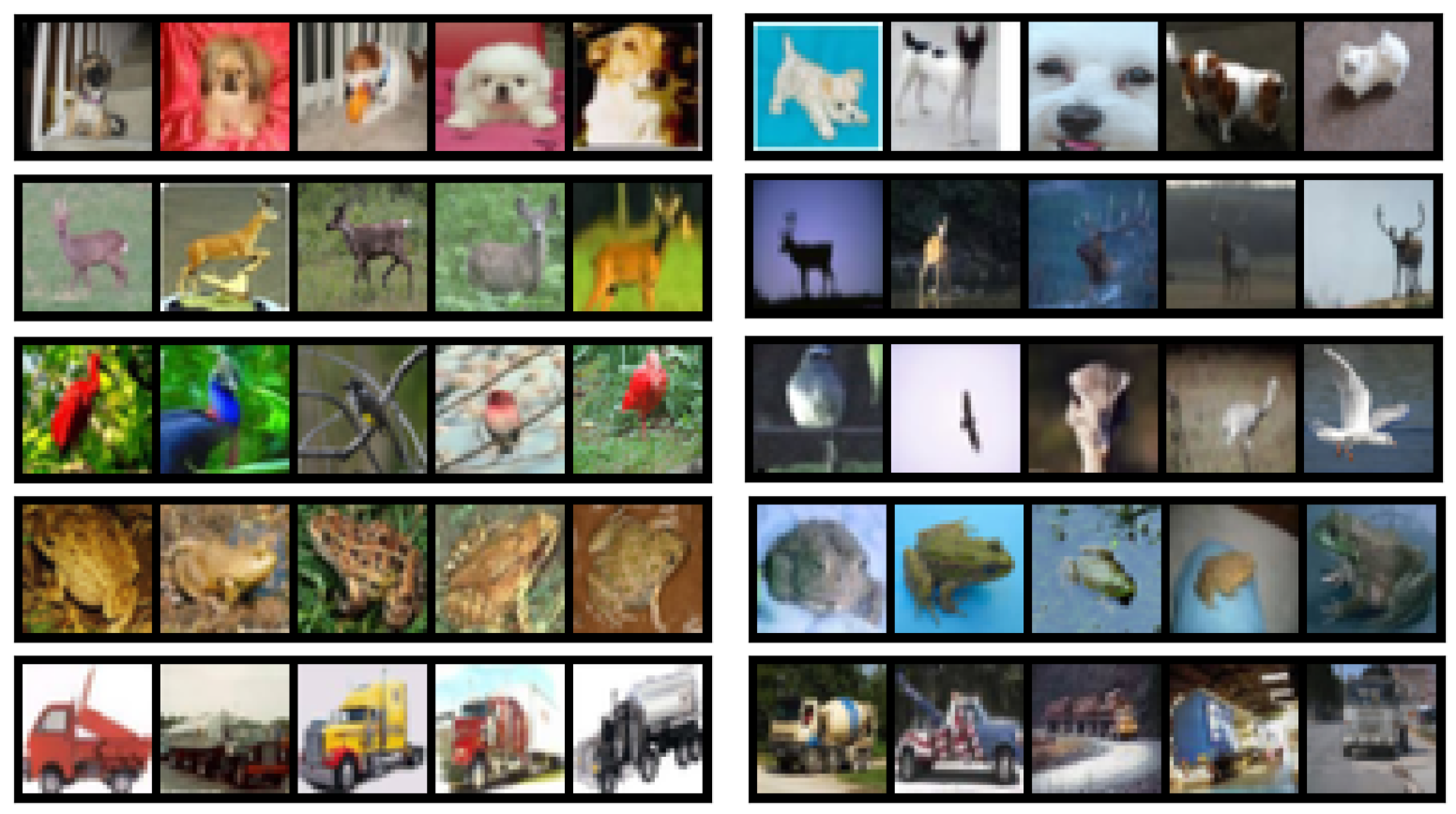}
      \centering
      \label{inlier_outlier_visualization_cifar10}
   }
   \caption{The top 5 most normal and most anomalous testing images for different classes. (a) Normal classes selected from MNIST: ``0'', ``1'', ``6'', ``7'' and ``9''. (b) Normal classes selected from CIFAR-10: ``Dog'', ``Deer'', ``Bird'', ``Frog'' and ``Truck''.}%
   \label{inlier_outlier_visualization}%
\end{figure}

To visualize both the normal class latent space and the reconstruction image space generated by DDR-ID, we use t-SNE visualization~\cite{maaten2008visualizing} to plot 2D embeddings of 500 randomly drawn testing samples in both spaces. For MNIST, we select class ``5'' to be the normal class and class ``ship'' for CIFAR-10. The DDR-ID model is trained using the normal class images and the class-specific reconstruction network $\mathcal{R}_{C}$ is conducted to generate high-dimensional representations for the 500 testing images. Fig.~\ref{t-sne-visualization} shows the t-SNE plots for both cases. It is observed from the 2D embeddings that the normal class testing images are clustered together centering at two templates $Z_C$ and $\mathcal{D}_{C}(Z_C)$, as expected. By sorting some testing images with respect to their anomaly scores, we could scrutinize them in both spaces. For example, as shown in Fig.~\ref{t-sne-5}, it is first observed that the normal class image with very low anomaly score (shown in the green rectangle) matches with the normal ``5'' from visual observation. Secondly, the anomalous class image with low anomaly score (shown in the red rectangle) appears to be a ``3'' which looks similar to ``5''. For the normal class image which deviates from the normal class distribution (shown in yellow rectangle), compared with other two anomalous digits ``9'' and ``8'' (shown below the yellow rectangle), it has lower $AS_l$ but higher $AS_{r}$. A similar analysis can be made for Fig.~\ref{t-sne-ship}. To summarize, the proposed anomaly scores can effectively quantify the anomalous degree for an unseen image while different anomaly scores might disagree with certain cases in the anomalous degree. Fig.~\ref{inlier_outlier_visualization} shows the samples of 5 most normal and 5 most anomalous testing images of normal classes selected from both datasets. {\color{black}{We refer to Section 2 of the supplementary material for the most normal and anomalous testing images of ALOCC~\cite{sabokrou2018adversarially} and Deep SVDD~\cite{ruff2018deep}.}}


{\color{black}{\subsection{Anomaly Detection on Endosome dataset}

In the section, we evaluate our method on the task of anomaly detection of biomedical patterns. We adopt Endosome dataset containing microscopic images of endosomes and non-endosome patterns~\cite{lin2019two}. Ring-like Endosomes are a type of organelles served as critical transport compartments that shuttle multiple nutrients inside cells. Hence, it is valuable to identify endosomes from non-endosome patterns. In this experiment, we set endosome patterns as the normal class and non-endosome patterns as the anomalous class. In this task, we train all the compared methods using only endosome patterns (normal class). The Endosome dataset consists of a training set and a testing set. The training set contains 165 endosome patterns and the testing set contains 214 endosome patterns and 366 non-endosome patterns. Fig.~\ref{EndosomeData} shows some examples of ring-like endosome patterns. We adopt the experiment protocol in ~\cite{ruff2018deep} and global contrast normalization for preprocessing. The experiment settings for the benchmarking methods are the same as specified in Section \ref{Settings}.

\begin{figure}[hptb]
   \centering
   \includegraphics[scale=1]{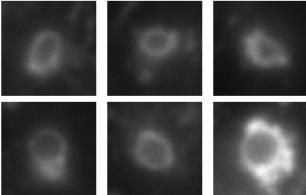}
   \caption{The illustration of the examples of ring-like endosome patterns in microscopic images.}
\label{EndosomeData}
\end{figure}

\begin{table}[htbp]\small
  \centering
  \caption{Average AUC in \% per method over 10 rounds of experiments on Endosome dataset. The best AUC value is in bold.}
    \begin{tabular}{|cccc|}
    \toprule
    \multicolumn{2}{|c}{METHOD} & \multicolumn{2}{c|}{AUC} \\
    \midrule
    \multicolumn{2}{|c}{ALOCC~\cite{sabokrou2018adversarially}} & \multicolumn{2}{c|}{76.0} \\
    \multicolumn{2}{|c}{DCAE~\cite{ruff2018deep}} & \multicolumn{2}{c|}{63.1} \\
    \multicolumn{2}{|c}{SOFT-BOUND DEEP SVDD~\cite{ruff2018deep}} & \multicolumn{2}{c|}{76.7} \\
    \multicolumn{2}{|c}{ONE-CLASS DEEP SVDD~\cite{ruff2018deep}} & \multicolumn{2}{c|}{78.0} \\
    \multicolumn{2}{|c}{DDR-ID} & \multicolumn{2}{c|}{\bf{79.1}} \\
    \bottomrule
    \end{tabular}%
  \label{EndosomeAUC}%
\end{table}%

\begin{figure}[hptb]
   \centering
   \subfloat[][]{
      \includegraphics[scale=0.5]{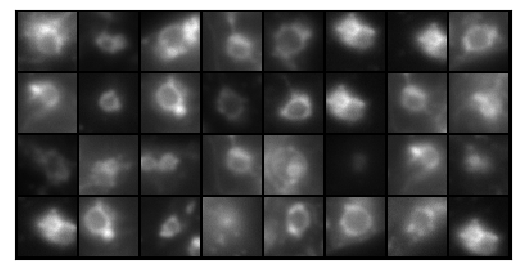}
      \centering
      \label{Endosome_normal}
       }
   \subfloat[][]{
      \includegraphics[scale=0.5]{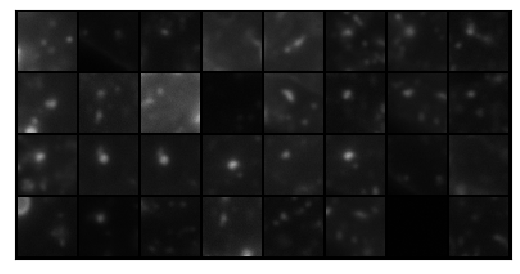}
      \centering
      \label{Endosome_abnormal}
   }
   \caption{The top 32 most normal and most anomalous testing images for Endosome dataset: (a) normal images; (b) anomalous images.}%
   \label{EndosomeQualitative}%
\end{figure}

From Table \ref{EndosomeAUC}, it is observed that our DDR-ID outperforms the competing methods in average AUC. Qualitatively, Fig.~\ref{EndosomeQualitative} shows 32 most normal and anomalous images measured by the anomaly scores. It is observed that the anomaly scores can reflect the degree of anomaly for endosome patterns. To summarize, the proposed DDR-ID is effective in anomaly detection of ring-like endosome patterns from non-endosome patterns in microscopic images.}}

\subsection{Detection of Adversarial Attacks on GTSRB Datasets}

\begin{table}[h]\small
  \centering
  \caption{Average AUC in \% per method over 10 rounds of experiments on GTSRB dataset. The best AUC value is in bold.}
    \begin{tabular}{cccc}
    \toprule
    \multicolumn{2}{c}{METHOD} & \multicolumn{2}{c}{AUC} \\
    \midrule
    \multicolumn{2}{c}{OC-SVM/SVDD~\cite{Tax2004}} & \multicolumn{2}{c}{67.5} \\
    \multicolumn{2}{c}{KDE~\cite{10.2307/2237880}} & \multicolumn{2}{c}{60.5} \\
    \multicolumn{2}{c}{IF~\cite{liu2008isolation}} & \multicolumn{2}{c}{73.8} \\
    \multicolumn{2}{c}{AnoGAN~\cite{schlegl2017unsupervised}} & \multicolumn{2}{c}{-} \\
    \multicolumn{2}{c}{DCAE~\cite{ruff2018deep}} & \multicolumn{2}{c}{79.1} \\
    \multicolumn{2}{c}{SOFT-BOUND DEEP SVDD~\cite{ruff2018deep}} & \multicolumn{2}{c}{77.8} \\
    \multicolumn{2}{c}{ONE-CLASS DEEP SVDD~\cite{ruff2018deep}} & \multicolumn{2}{c}{80.3} \\
    \multicolumn{2}{c}{DDR-ID} & \multicolumn{2}{c}{\textbf{83.2}} \\
    \bottomrule
    \end{tabular}%
  \label{AdversarialAUC}%
\end{table}%
\begin{figure}[h]
   \centering
   \subfloat[][]{
      \includegraphics[scale=0.4]{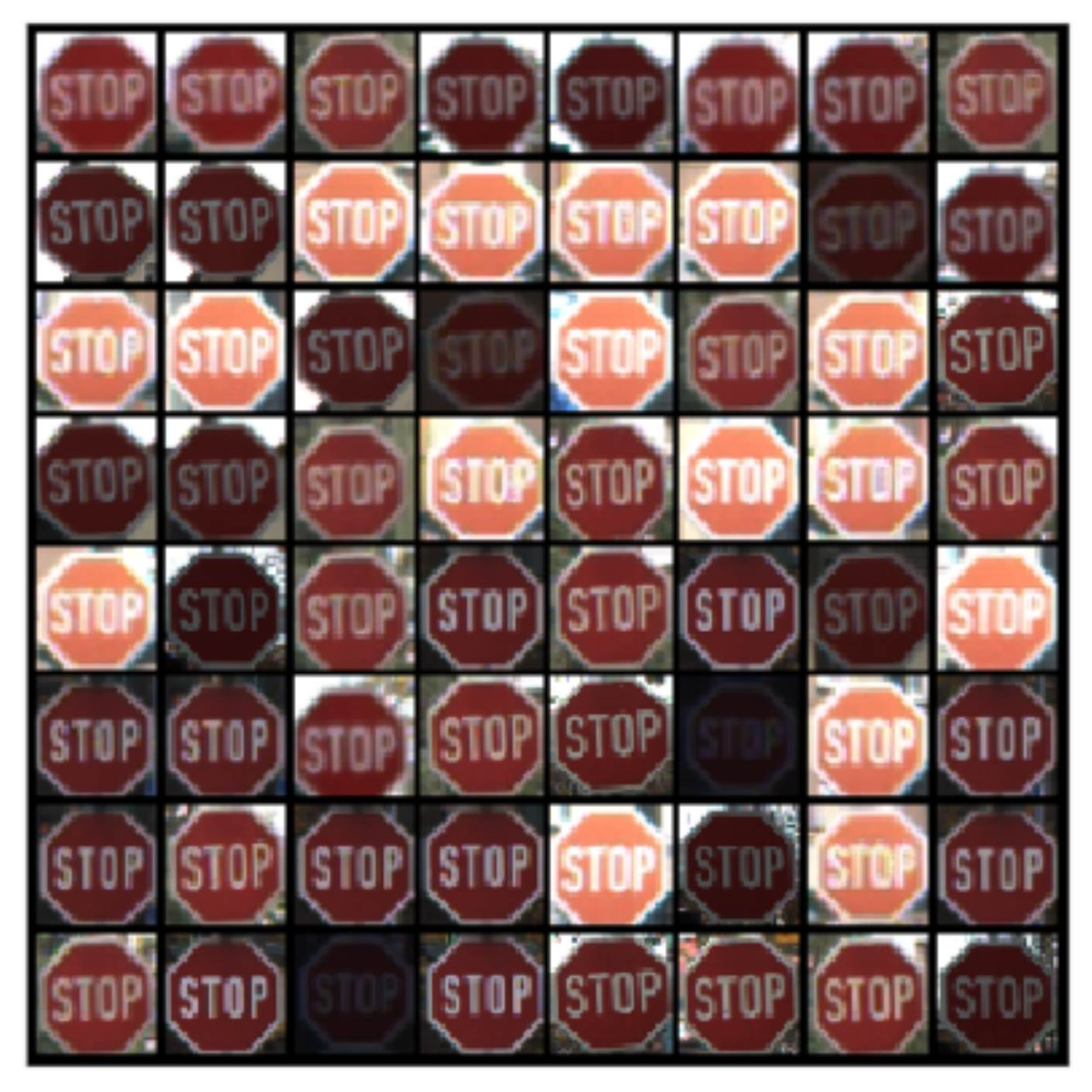}
      \centering
      \label{GTRSB_normal}
   }
   \subfloat[][]{
      \includegraphics[scale=0.4]{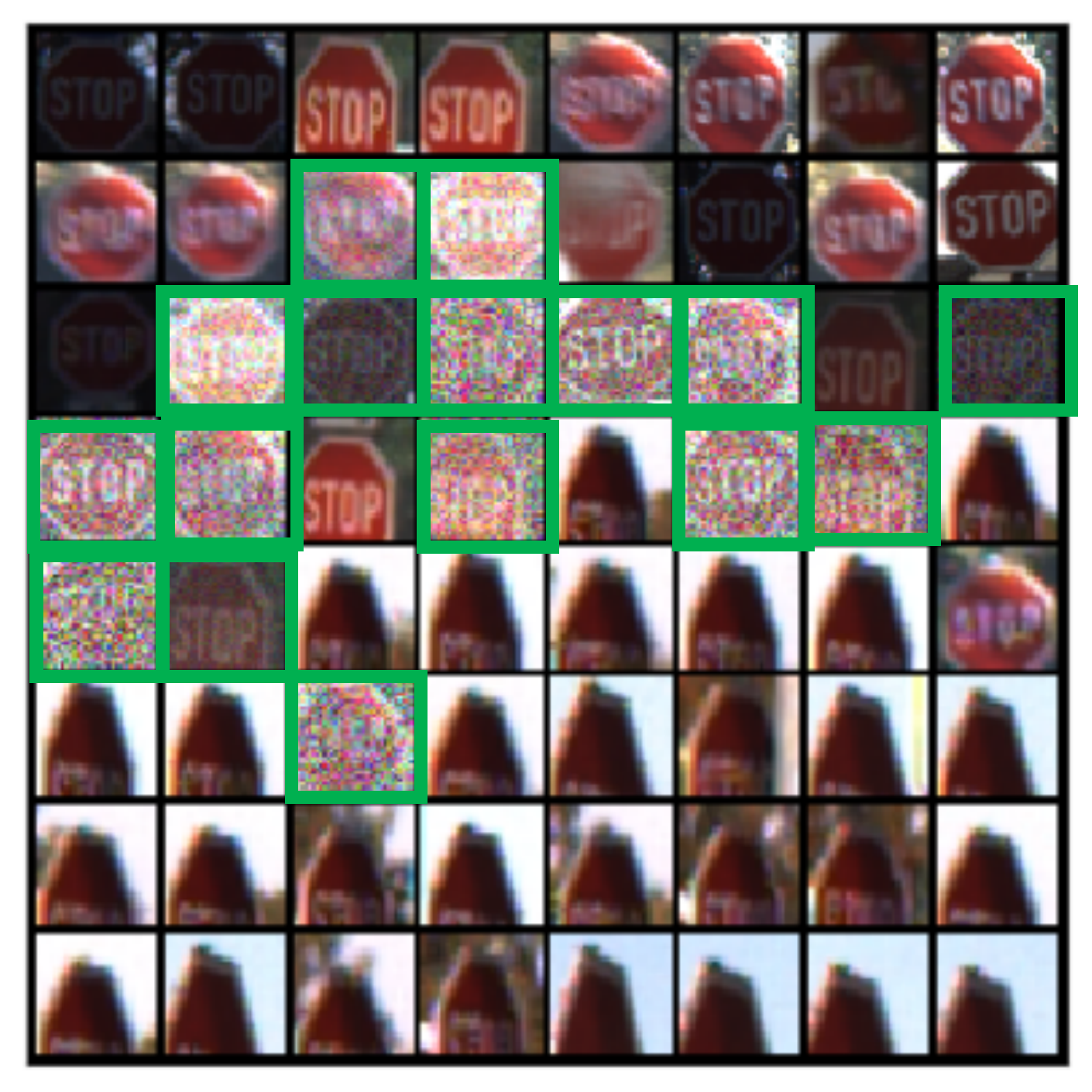}
      \centering
      \label{GTRSB_abnormal}
   }
   \caption{The top 64 most normal and most anomalous testing images for GTRSB: (a) normal images; (b) anomalous images where the adversarial samples are highlighted in the green rectangles. Best viewed in the color version.}%
   \label{AdversarialExamples}%
\end{figure}

In this section, we evaluate the performance of our method in detecting adversarial samples from normal ones. Following~\cite{ruff2018deep}, we use the ``stop sign'' class of the German Traffic Sign Recognition Benchmark (GTSRB) as the normal class and generate adversarial samples by randomly drawing stop sign images from the testing set and performing Boundary Attack proposed in~\cite{brendel2018decisionbased}. The training set has 780 images of stop signs and the testing set contains 270 normal samples and 20 adversarial samples. The 10\% border around each sign is removed. The same global contrast normalization applied in subsection \ref{MNISTandCIFAREx} is conducted for preprocessing. DDR-ID conducts 100 pretraining and 450 finetuning epochs, respectively. We compare our method with the results reported in ~\cite{ruff2018deep} on the average AUCs from 10 rounds of experiments.

Table \ref{AdversarialAUC} records the AUC results. It is noted that AnoGAN did not converge in this dataset due to small data size which is not sufficient for GANs. It is observed that our DDR-ID significantly outperforms all the competing methods in average AUC. Qualitatively, Fig.~\ref{AdversarialExamples} shows 64 most normal and anomalous images measured by the anomaly scores. From Fig.~\ref{GTRSB_normal}, it can be concluded that the detection of DDR-ID is robust with respect to illumination conditions as the normal images contain exact stops signs under various lighting conditions. From Fig.~\ref{GTRSB_abnormal}, it is noted the anomalous images consist of the adversarial samples, the stop signs taken from bad camera angles and some blurred images. This shows that our DDR-ID could not only robustly identify normal samples in various scenarios but also detect different types of anomaly which are not accessible during training. To summarize, the proposed DDR-ID is capable of effective detection of image adversarial attacks.

\section{Ablation study on the anomaly scores}
{\color{black}{
In this section, we conduct an ablation study on the effect of different anomaly scores. We mainly consider four ablation methods: (i) DDR-ID($AS_l$), i.e., DDR-ID with anomaly score $AS_l$ defined in Eq.~(\ref{AS_l}); (ii) DDR-ID($AS_r$), i.e., DDR-ID with anomaly score $AS_r$ defined in Eq.~(\ref{AS_r}); (iii) DDR-ID($AS_l + AS_r$), i.e., DDR-ID with anomaly score defined as $AS_l + AS_r$; (iv) DDR-ID (val), i.e., DDR-ID with anomaly score determined using a hold-out validation set. We conduct the ablation experiments using both MNIST and CIFAR-10 datasets.

Table~\ref{tab:AUCs_ablation} records the average AUCs for all the ablation methods. On one hand, DDR-ID (val) consistently outperforms DDR-ID($AS_l$), DDR-ID($AS_r$) and DDR-ID($AS_l + AS_r$) on two datasets. On the other hand, compared with DDR-ID($AS_l$) and DDR-ID($AS_s$), DDR-ID($AS_l + AS_r$) produces a slightly better result on MNIST but a deteriorated performance on CIFAR-10. Therefore, we conclude that a simple summation of $AS_l$ and $AS_r$ can not generate a consistently good anomaly score. Hence, we adopt DDR-ID (val). Specifically, for MNIST, our method adopts $AS_l$ for 6 classes and $AS_r$ for the remaining 4 classes. For CIFAR-10, our method adopts $AS_l$ for 7 classes and $AS_r$ for the remaining 3 classes.

\begin{table}[htpb]\small
  \centering
  \caption{Average AUCs in \% over 10 rounds of experiments for ablation study. The best AUC value is in bold.}
    \begin{tabular}{|p{15.5em}|c|c|}
    \toprule
    \multicolumn{1}{|c|}{\multirow{2}[3]{*}{Method}} & \multicolumn{2}{c|}{Datasets} \\
\cmidrule{2-3}    \multicolumn{1}{|c|}{} & MNIST & CIFAR-10 \\
\midrule
    \multicolumn{1}{|c|}{DDR-ID($AS_l$)} & 95.6  & 63.7 \\
    \multicolumn{1}{|c|}{DDR-ID($AS_r$)} & 95.7  & 62.4 \\
    \multicolumn{1}{|c|}{DDR-ID($AS_l + AS_r$)} & 95.8  & 60.5 \\
    \multicolumn{1}{|c|}{DDR-ID (val)} & \textbf{96.2}  & \textbf{65.4} \\
    \bottomrule
    \end{tabular}%
  \label{tab:AUCs_ablation}%
\end{table}%

}}

\section{Conclusion}
In this paper, we have proposed an image anomaly detection (AD) method called dual deep reconstruction networks based image decomposition (DDR-ID). Our network extracts normal class information by decomposing an image into its class-specific component and non-class-specific component. Such decomposition network is trained using an end-to-end learning process which jointly optimizes for three losses: the one-class loss, the auxiliary latent space constrain loss and the reconstruction loss. Based on the normal-class-specific component, we define two anomaly scores to quantify the anomalous degree of a testing image in either normal class latent space or reconstruction image space. Thereby, effective anomaly detection can be performed via thresholding the anomaly score. Our experiments demonstrate that, quantitatively and qualitatively, our DDR-ID outperforms multiple related benchmarking methods in image anomaly detection using MNIST, CIFAR-10 and Endosome datasets and adversarial attack detection using GTSRB dataset.

\section{Conflict of Interest}
The authors declared that there is no conflict of interest.





{\small
\bibliographystyle{model1-num-names.bst}
\bibliography{OurRef}
}




\end{document}